\RequirePackage{fix-cm}
\RequirePackage{amsmath}
\documentclass[twocolumn]{svjour3}          
\smartqed  

\usepackage[numbers]{natbib}

\usepackage{amsmath,amssymb,amsfonts}
\usepackage{color,soul}
\usepackage{graphicx}
\usepackage{bm}

\usepackage{booktabs} 
\usepackage{mathtools}
\usepackage{algorithm,algorithmic}
\usepackage{tikz}
\usepackage{float}
\usepackage{wrapfig}
\usepackage{makecell}
\usepackage{amsmath}
\makeatletter
\renewcommand*\env@matrix[1][*\c@MaxMatrixCols c]{%
  \hskip -\arraycolsep
  \let\@ifnextchar\new@ifnextchar
  \array{#1}}
\makeatother

\begin{document}


\title{WisdomNet: Trustable Machine Learning Towards Error-Free Classification}                      




\author{Truong X. Tran \and Ramazan S. Aygun}

\institute{Truong X. Tran and Ramazan S. Aygun \at
              Computer Science Department, The University of Alabama in Huntsville, USA.
              \email{\{txt0002,aygunr\}@uah.edu}           
}
\date{}

\maketitle

\begin{abstract}
Misclassification is a critical problem in many machine learning applications. 
Since even the classifier models with high accuracy (e.g., $>95\%$) still introduce some misclassification error, it may not be possible to rely on the output of a classifier. 
In this paper, we introduce {\em trustable learning,} which prompts the learning model to yield only the true output, thus avoiding misclassifications. Whenever the model cannot decide the output accurately, the learning model should indicate that there could be a misclassification error if it is forced to classify, and hence it should reject to make a decision or defer it to a human expert. Therefore, we develop a methodology for trustable learning and apply it to artificial neural networks and show that it is possible to develop a classifier with $0\%$ misclassification error. 
We propose a novel neural network architecture named {\em WisdomNet} that could provide zero prediction error by introducing an additional neuron named as {\em conjugate neuron} that would indicate whether the network is able to classify the data correctly or not.
The WisdomNet architecture can be applied to any previously built model, and we have evaluated WisdomNet with several network architectures such as multilayer perceptron, convolutional neural network, and deep network on different data sets. The results show that the WisdomNet is able to reduce the classification error rate to $0\%$ while labeling the data difficult to classify as `reject' at a low percentage of within around $10\%$.
\end{abstract}





\section{Introduction}
The quality of learning algorithms is crucial in many applications. For the real-world applications such as medical diagnosis \cite{valdovinos2005medicalapp}, disaster prediction \cite{mandal2005neural}, fraud detection \cite{fawcett1997adaptive}, etc. where misclassifications are costly, a single prediction error may yield information leakage, fatal, disastrous and costly outcomes. 
It is common to evaluate the performance of classification models based on some evaluation metrics, such as accuracy. While some models integrate the cost of misclassification along with the accuracy for choosing the best model, these models can still yield incorrect predictions.
Alpaydin suggests that a learning model may {\em reject} to make a decision or defer the decision to a human expert if the human expert is likely to make a better decision than the learning model, especially for cases where wrong decisions could be costly or fatal \cite{Alpaydin2014IML}.

In many applications, high accuracy such as higher than $95\%$ may not satisfy the experts since incorrect predictions can still occur. For example, diagnosing cancer, detecting a tornado, identifying fraud, determining an attack, classifying a protein crystal, etc. are some examples where false negatives are not acceptable at all. Including misclassification cost may help choose models with fewer false negatives and higher false positives, but may not eliminate the classification error.
Binary classification is useless for determining the misclassified data as the expert should go through all negative samples to determine false negatives and also positive samples since they are the desired outputs. This process implies the manual scan of all data by the expert and makes the classifier useless.  
In our previous researches on protein crystallization analysis, our crystallographer recommended having one more class between crystal and non-crystal categories, named as likely-leads \cite{pusey2017:data, Subedi2020VisualX2, Tran2017:subordinate, tran2020protein}. In this circumstance, the missed crystals hopefully could be classified as likely-leads rather than non-crystals, and the expert could review experiments labeled as likely-leads manually to avoid any missed crystals without requiring to review non-crystals.

The major issue is that whenever specialists work on sensitive data, they cannot fully trust the artificial learning model. 
To prevent false predictions, experts manually review the results. However, the manual scanning of prediction results voids the benefits of computer-assisted learning models. In this paper, our goal is to develop {\em trustable learning} method, which prompts the learning model to yield only the true output, thus avoiding misclassifications. Our method does not require the creation of another class and labeling of such data (e.g., the likely-leads category in protein crystallization research). For example, in hand-writing character recognition, the system should not require the labeling of confusing examples between `B' and `8'. From the training data, the trustable learning model should be able to determine what it cannot classify properly and reject to make a decision.
This trust problem cannot be solved by minimizing the total error or loss function as in neural networks. When the accuracy becomes too high, overfitting is likely to be an issue. The trustable learning model should be free of the overfitting problem.
In addition, we should note that our trustable learning method is not a multiclass classification problem of adding one more class for the data samples that are hard to classify. Firstly, classification errors can still occur for multiclass problems, and actually, there will be more combinations of misclassifications. But, our trustable learning method aims to eliminate classification errors. Secondly, the data will be unbalanced when the number of data samples that are hard to classify is small, and it will be difficult to train a multinomial classifier to distinguish these problematic instances.

In this paper, we apply our trustable learning methodology to artificial neural networks. We propose a novel neural network architecture named {\em WisdomNet} that could provide zero prediction error by introducing an additional neuron named as {\em conjugate neuron} that would indicate whether the network can classify the data correctly or not. 

Firstly, a WisdomNet network is created from a trained neural network. The WisdomNet adds to the output layer a conjugate neuron that can learn the doubtful data, which might incorrectly be classified by the original base network. Then, the WisdomNet initially keeps all weights and biases that are inherited from the base network but initializes the relevant parameters of the conjugate neuron with zeros. Finally, the WisdomNet is trained only on the data items that were misclassified by the base network. The WisdomNet inherits the knowledge learned by the base network and then learns more from the doubtful data instances. 
The WisdomNet architecture can be applied to any previously built model, and we have evaluated WisdomNet with several network architectures such as multilayer perceptron, convolutional neural network, and deep network on different data sets. Our goal in this paper is trustable learning for binary classification. The results show that the WisdomNet is able to reduce the classification error rate to $0\%$ while labeling the data doubtful to classify as `reject' (i.e., reject to make a decision) at a low percentage of within around 10\%. We also provide a discussion on how trustable learning can be applied to multi-class classification problems.

The contributions of our paper are summarized as follows:
\begin{enumerate}
	\item {Trustable learning method that can reduce the misclassification rate to $0\%$,}
	\item {A new architecture named as the WisdomNet for trustable learning of artificial neural networks,}
	\item {Application of the WisdomNet on various learning architectures including multi-layer perceptron and deep learning architectures, and the reduction of the classification error to $0\%$ while keeping the `reject' within around $10\%$,}
	\item {Evaluation of the main factors that affect the performance of trustable learning (the WisdomNet in this context).}
\end{enumerate}

The rest of the paper is organized as follows. The following section provides the related work. The methodology of trustable learning and the WisdomNet is described in Section \ref{propose}. Experiments and results are provided in Section \ref{experiment}. A discussion on the extension of the WisdomNet is presented in Section \ref{discussion}. The last section provides the conclusion of the paper.

\section{Related work} \label{relatedwork}
In the last decade, neural networks with deep learning architectures boosted the performance of many machine learning problems such as image recognition \cite{milletari2016vnet}, speech recognition \cite{deng2013new}, natural language processing \cite{kumar2016ask,sarikaya2014application}, market prediction \cite{li2005forecasting}, medical diagnosis \cite{liu2014early}, cybersecurity \cite{buczak2016survey}, etc. Many of these applications aim to achieve or surpass human level performance by allowing errors at a level.
Their main goal is typically to reduce the classification error. In some applications such as object recognition, it may not be a critical mistake to mislabel a dog as a cat, but it could be a fatal error to mislabel a friend as an enemy target or to miss cancer cells or tumors. In supervised learning, research studies consider problems related to bias/variance or overfitting/underfitting. Hence, they aim to minimize the classification error on training data while targeting high accuracy on test data. 

The cases of doubt where a classifier may need to {\em reject} to make a decision or defer the decision to an expert have been mentioned by Alpaydin \cite{Alpaydin2014IML}. 
For example, in medical diagnosis, since an incorrect decision may yield to improper treatment, in such a case avoiding or rejecting a decision or transferring the decision to an expert may be a better option. The author states {\em reject} as a choice between computer based decision and human expert decision when the computer program error is more likely than the error of a human expert. Alpaydin covers several scenarios for cases of doubt where a classifier may reject to make a decision. For example, for a classification problem, it is possible to visualize the doubt region in a given feature space by the most specific hypothesis and the most general hypothesis. The most specific hypothesis, $S$, includes all observed positive samples without any negative cases and covers as small feature space as possible. The most general hypothesis, $G$, includes all observed positive examples and covers as much of the feature space without containing any negative examples. Any sample that is not covered by $S$ but included by $G$ is considered as a case of {\em doubt}, and the system may reject the classification to delegate to a human expert. For multi-class learning problems, if multiple hypotheses return true for their classes, the class of the sample cannot be chosen, and this case is again stated as a case of doubt and should be rejected. For Bayesian classifier, a sample is rejected if $P(C_i|\textbf{x})\leq1-\lambda$ even if $P(C_i|\textbf{x})$ is greater than the probability of $\textbf{x}$ belonging to any other class (where $\lambda$ is defined as loss/cost for choosing an action corresponding to reject). Another case for rejection is for discriminant classifiers of multiple classes where multiple positive half-spaces of hyperplanes overlap. These scenarios are reasonable scenarios where a classifier may opt to reject to make a decision. However, it is also important to quantify the ratio of rejections and understand the tradeoff between reject and misclassifications. In this paper, we focus on rejection to reduce the misclassification error to zero.
Rather than just stating {\em reject}, our method also aims to reduce the error to zero while keeping the ratio of reject cases as minimum as possible. Moreover, the application of reject scenarios is not clear for neural networks and not mentioned in \cite{Alpaydin2014IML}. 
In this paper, we also show how to reject scenarios can be determined for neural networks.

In the literature, there has been recent work that mentions trustable or trustworthy machine learning. Seshia et al. \cite{seshia2016design} state three properties for trustworthy machine learning for cyber-physical systems that require verification of system correctness: i) algorithms should be explanatory in addition to being predictive, ii) the training and testing data should be systematically generated to yield formal guarantees to converge to the target goal, and iii) there must be a specification for what correctness actually means. Similarly, Kaul \cite{kaul2018speed} mentions three properties for trustworthy machine learning: i) robustness against inconsistent or corrupted training data, ii) resilience to adversarial manipulation of test data, and iii) accountability, transparency and/or fairness of the predictions. On the other hand, Varshney  \cite{varshney2019trustworthy} lists six items for trustworthy machine learning: i) robustness to data shift (covariate shift, prior probability shift, concept drift, and confounding shift), ii) protection from data poisoning (injection of carefully designed data), iii) fairness (avoiding biases and prejudices), iv) interpretability, v) system-level transparency (clear specification of how a model can be used, tested and standardized), and vi) social good. In a special issue on trustable machine learning in \cite{Editorial2018}, the robustness and interpretability of learning algorithms are stated as the driving factors for gaining the trust of patients as well as health care providers. These factors can be considered as the factors that influence the trust of users.

If the predictions of a machine learning algorithm are interpretable, experts or users make sense of predictions and avoid crucial mistakes by identifying possible mispredictions based on explanations. As an example of machine learning that supports interpretability, Lundberg et al. \cite{lundberg2018explainable} aim at providing explanations of predictions during surgery to prevent hypoxaemia, thus helping increase the anticipation of events by anesthesiologists from $15\%$ to $30\%$ using trained gradient boosting machines. Teso and Kersting \cite{teso2019explanatory} find out that explanatory interactive learning, where the machine learning algorithm gets feedback from the user about its prediction, can encourage trust into the model. Their method combines active learning with local explainers.   Moreover, it may be possible to increase the trust of the user by providing information about how the model behaves based on some inputs. Zhou et al. \cite{zhou2019physiological} show that presenting the influence of training data using visualization based on parallel coordinates significantly increases user trust. However, increasing the trust of the user may not make the model correct. Moreover, it is hard to interpret the models based on neural networks, especially deep learning architectures.

Another issue that affects the trust in machine learning algorithms is the improper training of learning models due to the quality of data. The quality of data may not only affect the accuracy of the model, but it may also generate biased or unfair predictions. Zhang et al. \cite{zhang2018training} aim at finding problematic or incorrectly labeled data items by using a small set of data items whose truthness is verified. Similarly, Han et al. \cite{han2019robust} develop federated model training with collaborative learning agents sharing information about data tuning using trusted instances. Robustness to possible model attacks is another issue about trust. Jun et al. \cite{jun2018adversarial} define trustworthy machine learning as machine learning systems that can determine how they can be attacked in the context of adversarial attacks and propose a solution for stochastic multi-armed bandits that are vulnerable to reward manipulation. Chen et al. \cite{chen2019robust} propose robust attribution regularization based on integrated gradients to study how the output of one-layer neural networks may change due to the input changes. 

An interesting method that provides feedback to the expert is proposed by Wang et al. \cite{wang2018development} for polyp detection that achieves $94.38\%$ per-image-specificity on colonoscopy images set, $91.64\%$ per-image-sensitivity on colonoscopy videos, and $95.40\%$ per-image-specificity on unaltered full-range colonoscopy videos without polyps. They trained an algorithm based on local features for polyps appearing partially. These partial polyps were partially tagged, and they were reported as well to avoid missing polyps. Their work gave a chance for the expert to go over partially tagged items.

We should note that our trustable learning methodology is not similar to the multi-class classification problem. For example, in speech recognition application, the keyword spotting (e.g., ``Hi Siri" or ``Ok Google") is quite important to start the query with smart devices \cite{team2017hey}. Since it is likely that the user may utter words that are not in the training set, the learning model still needs to determine those unrecognized words. To train the network for ignoring irrelevant sounds, some samples that are not in the target classes should be provided. To achieve this, some applications add data items with the ``unknown" class label to the training set. For example,  the Google Speech Command data set contains a special label as ``unknown word" among other speech commands \cite{warden2018speech}. In this case, the ``unknown" class is one of the categories for training the model.  Our training method for trustable learning is not a multiclass classification training process where misclassifications between any pairs of classes can still occur. Moreover, the multiclass classification will suffer imbalance since the number of misclassified samples will be few for a learning model with high accuracy.

In deep learning, it is popular to use transfer learning \cite{transferpan2009, yann2016Deeplearning, lecun2015deep} to adapt the model that is trained on a large data set for using it in another domain. It is also popular to transfer all parameters of a base network, like in fine-tuning, so that the new model inherits all the learned features and just learns to classify for the new domain. Our method is different from transfer learning since we still work on the same domain and use the same data. Moreover, our method is not reinforcement learning, where the learning model learns by trial and error.

In the literature, to the best of our knowledge, we were not able to find research studies aiming at the development and implementation of zero error learning models. This could be due to the overfitting issue. Recently, we proposed {\em else-tree} to reduce the error to zero percent using the decision tree classifier \cite{tran2018else}. 
The else-tree minimizes the error by not assigning a class label to the sample that could be misclassified and follows the else-branch in the tree. The else-tree generates a virtual class named $undecided$ that did not exist in the original data. It was able to reach $0\%$ error for some data sets by moving the hard to classify data into the $undecided$ class. 

Some of the ideas on trustworthy machine learning are conceptual, while there are concrete methods to improve trust, such as improving the quality of data. Since the interpretability of the model is domain-specific, it could be a barrier to the generalizability of learning models. Meanwhile, models based on neural networks are hardly interpretable, and this might imply that it may not be possible to develop trustable neural network architectures. In \cite{CACMStaff2019}, there is an argument that a neural network may not only fail when the network encounters data radically different from the training data, but it may even fail even if the new data is very similar to the training data as in the case of driverless cars \cite{Zhou2019MTD}. The question of trust becomes harder even if the data similar to the training data cannot be classified properly. 
We have a supposition of why learning models fail on data similar to the training data. We conjecture that some of the correct classifications of a high accurate learning model could be just `lucky' predictions. While improving the accuracy of learning models, experts typically analyze incorrect classifications while ignoring lucky predictions.  
Our approach is based on whether the model can determine what it cannot learn. Hopefully, our trustable learning method will not only learn data that would normally yield incorrect classifications but also determine the ones that are luckily correctly classified. Our goal is to build models that can reduce the error to 0\% without overfitting.

\tikzset{%
   neuron missing/.style={
    draw=none, 
    scale=2.5,
    text height=0.333cm,
    execute at begin node=\color{black}$\vdots$
  },
}

\begin{figure*}[htb]
\begin{tikzpicture}[x=0.8cm, y=0.8cm, >=stealth]

\foreach \m/\l [count=\y] in {1,2}
{
 \node [circle,fill=gray!50,minimum size=0.5cm] (input-\m) at (0,2.5-\y) {};
}
\foreach \m/\l [count=\y] in {3}
{
 \node [circle,fill=gray!50,minimum size=0.5cm ] (input-\m) at (0,-1.5) {};
}
 
 \node [neuron missing]  at (0,-0.5) {};

\foreach \m/\l [count=\y] in {1,2}
{
 \node [circle,fill=gray!50,minimum size=0.5cm] (hidden-\m) at (2,2.5-\y) {};
}
  
\foreach \m [count=\y] in {3}
  \node [circle,fill=gray!50,minimum size=0.5cm ] (hidden-\m) at (2,-1.55) {};
  
 \node [neuron missing]  at (2,-0.5) {};

\foreach \m/\l [count=\y] in {1,2}
{
 \node [circle,fill=gray!50,minimum size=0.5cm] (hidden1-\m) at (4,2.5-\y) {};
}
  
\foreach \m [count=\y] in {3}
  \node [circle,fill=gray!50,minimum size=0.5cm ] (hidden1-\m) at (4,-1.55) {};
  
 \node [neuron missing]  at (4,-0.5) {};

\foreach \m [count=\y] in {1,2}
  \node [circle,fill=gray!50,minimum size=0.5cm ] (output-\m) at (6,2-\y) {};

\foreach \i in {1,...,3}
  \foreach \j in {1,...,3}
    \draw [->] (input-\i) -- (hidden-\j);

\foreach \i in {1,...,3}
  \foreach \j in {1,...,3}
    \draw [->] (hidden-\i) -- (hidden1-\j);
    
\foreach \i in {1,...,3}
  \foreach \j in {1,2}
    \draw [->] (hidden1-\i) -- (output-\j);
    
\node [above] at (0,2) {Input layer};
\node [above] at (3,2) {Hidden layers};
\node [above] at (6,2) {Output layer};

\node [right] at (output-1.east) {Output-0};
\node [right] at (output-2.east) {Output-1};


\foreach \m/\l [count=\y] in {1,2}
{
 \node [circle,fill=gray!50,minimum size=0.5cm] (vinput-\m) at (12,2.5-\y) {};
}
\foreach \m/\l [count=\y] in {3}
{
 \node [circle,fill=gray!50,minimum size=0.5cm ] (vinput-\m) at (12,-1.5) {};
}
 
 \node [neuron missing]  at (12,-0.5) {};

\foreach \m/\l [count=\y] in {1,2}
{
 \node [circle,fill=gray!50,minimum size=0.5cm] (vhidden-\m) at (14,2.5-\y) {};
}
  
\foreach \m [count=\y] in {3}
  \node [circle,fill=gray!50,minimum size=0.5cm ] (vhidden-\m) at (14,-1.55) {};
  
 \node [neuron missing]  at (14,-0.5) {};

\foreach \m/\l [count=\y] in {1,2}
{
 \node [circle,fill=gray!50,minimum size=0.5cm] (vhidden1-\m) at (16,2.5-\y) {};
}
  
\foreach \m [count=\y] in {3}
  \node [circle,fill=gray!50,minimum size=0.5cm ] (vhidden1-\m) at (16,-1.55) {};
  
 \node [neuron missing]  at (16,-0.5) {};

\foreach \m [count=\y] in {1,2}
  \node [circle,fill=gray!50,minimum size=0.5cm ] (voutput-\m) at (18,2-\y) {};
  
\foreach \m [count=\y] in {1}
  \node [circle,fill=brown,minimum size=0.5cm ] (vundecided-\m) at (18,-1) {};

\foreach \i in {1,...,3}
  \foreach \j in {1,...,3}
    \draw [->] (vinput-\i) -- (vhidden-\j);

\foreach \i in {1,...,3}
  \foreach \j in {1,...,3}
    \draw [->] (vhidden-\i) -- (vhidden1-\j);
    
\foreach \i in {1,...,3}
  \foreach \j in {1,2}
    \draw [->] (vhidden1-\i) -- (voutput-\j);
    
\foreach \i in {1,...,3}
  \foreach \j in {1}
    \draw [red,->] (vhidden1-\i) -- (vundecided-\j);
    
\node [above] at (12,2) {Input layer};
\node [above] at (15,2) {Hidden layers};
\node [above] at (18,2) {Output layer};

\node [right] at (voutput-1.east) {Decided output-0};
\node [right] at (voutput-2.east) {Decided output-1};
\node [right] at (vundecided-1.east) {Reject output};


\end{tikzpicture}
~~~~~~~~~~~~~~~~~~~~~~~~~~~~~~~~~~~~~~~~~~(a) Standard Neural Network ~~~~~~~~~~~~~~~~~~~~~~~~~~~~~~~~~~~~~~~~~~~~~~~~~~~~~~~~~~~~~~~~~~~~~~~~(b) WisdomNet Architecture
\caption{Example of WisdomNet Architecture. a) A standard neural network with 2 hidden layers; b) The WisdomNet created by applying the structure of the network in (a); The conjugate neuron (brown node) is added to the output layer and connected to the last hidden layer of the network (red arrows representing weights).}
\label{fig-WisdomNetmodel}
\end{figure*}
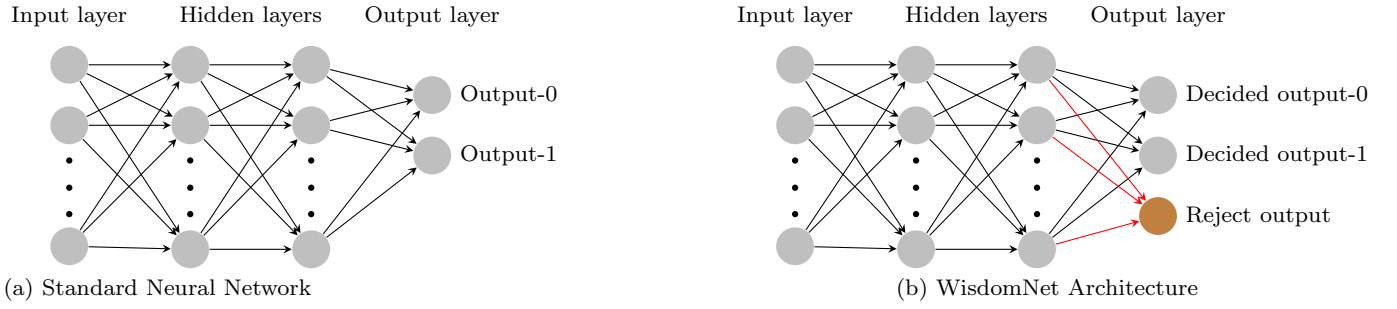

\section{Trustable Learning} \label{propose}
Trustable learning is a supervised learning method where the learning model can also determine what it cannot learn and yields `reject' in the likelihood of misclassification. Hence, trustable learning delivers learning models that can reduce the classification error to 0 by labeling data hard to classify as `reject to make a decision', `defer to a human expert', `undecided', `unable to classify', `ambiguity in classification', `doubtful', or `misclassification likely'. We will use the term `reject' \cite{Alpaydin2014IML} to cover all these types of cases to avoid misclassification. Throughout the paper, we use `reject' as a label to indicate the action of `reject to make a decision'. Moreover, we also use it as an adjective as in `reject' sample to mean `reject to make a decision for the sample' or `reject' set to indicate the set of samples that will be rejected to make a decision.  

{\bf Why to Reject to Make a Decision?} The users of a learning model should firstly consider whether misclassifications by a classifier, $C$, with an accuracy of $P\%$ and false negative rate of $F$ are tolerable or not. In applications such as medical diagnosis, weather forecasting, network security attacks, protein crystallization, or self-driving cars making wrong decisions yielding fatal or costly outcomes are not acceptable. In such cases, human expert opinion or interruption may be needed. For example, consider cases of where classifiers i) miss the presence of a tornado and cost the lives of people, ii) miss a malicious tumor and thus failing to notify an expert to start the treatment in time, iii) miss change of lanes due to road construction, iv) miss a network attack, or v) miss a crystallization condition that is crucial in the development of a drug. When traditional learning techniques are used, misclassifications are always likely to occur. In these critical applications, it would be desired if the classifier can defer the decision to a human expert so that tornadoes, malicious tumors, network attacks, lane changes, or protein crystals can be determined by an expert and necessary precautions can be taken. Nevertheless, delegating the decision to a human expert also has some costs, hence the number of rejections should be kept minimal.

{\bf Not to Be Fooled by Lucky Predictions.} One way to analyze the output of a classifier to improve its accuracy is to study its incorrect predictions. By analyzing those misclassifications, the learning model, the feature set, or the preprocessing of data can be optimized to yield better results. One fallacy of this reasoning is the assumption that correct predictions made by the learning model were true classifications by the model.  As mentioned in \cite{CACMStaff2019}, the learning model may fail even if the new data is similar to the training data, as in the case of driverless cars \cite{Zhou2019MTD}. Normally, we are able to observe the misclassifications by a classifier in the training phase, nevertheless, the lucky predictions which are likely to fail in real scenarios cannot be determined. It is desirable for trustable learning methods to learn these lucky cases as well.
If the learning model only learns misclassified samples, that would be considered as overfitting. The learning model should not only learn those instances but maybe learn other problematic instances that were classified correctly but could actually be classified incorrectly. 

In this paper, as an example of trustable learning, we develop a new architecture for neural models named as the WisdomNet. Our architecture is able to learn when to reject the decision to classify and can defer to a human expert. Meanwhile, it also aims to keep the ratio of rejected samples as low as possible.

\subsection{WisdomNet}

The WisdomNet is a neural network architecture for trustable learning that can determine what it cannot correctly classify. Its name is inspired by Aristotle's quote, ``Knowing yourself is the beginning of all wisdom." As mentioned in \cite{kim2019artificial}, artificial intelligence learning systems should have an ability to recognize their ignorance and show not merely intelligence but wisdom as well. We propose a simple but effective method to learn what a learning model cannot learn properly. 

In this paper, we focus on the application of the WisdomNet on binary classification problems. The WisdomNet is built in 4 steps: 1) Pre-train a standard neural network, 2) Identify the misclassified training samples, 3) Create a new network from the trained network and add a new {\em conjugate neuron} to the output layer for outputting `reject' or `unable to classify', and 4) Fine-train the new network with only the misclassified training samples.
The {\em conjugate neuron} is similar to the neuron outputting the class label, but in this case, it outputs `reject' or `unable to classify' instead of the original class label. The procedure of training the WisdomNet is summarized in Table \ref{tab:algol}. 

\begin{table}[t]
\caption{ The training process of the WisdomNet.}
\label{tab:algol}
\resizebox{\columnwidth}{!}{
\begin{tabular}{l}
\hline
\textbf{Pre-Training Phase: Train a standard $\mathcal{N}$ network}  \\
\begin{tabular}[c]{@{}l@{}} 
1. Train $\mathcal{N}$ using standard backpropagation and \\ stochastics gradient descent algorithm with the training set $S$.\end{tabular}                                                  \\

\textbf{Identify Misclassified Samples}\\
1. Classify the training set $S$ using the trained network $\mathcal{N}$.        \\
\begin{tabular}[c]{@{}l@{}}
2. Add all misclassified data samples to the reject set $S^r$  \\
3. Assign a new `reject' label to these samples.\end{tabular}                                                  \\

\textbf{Construct the WisdomNet $\mathcal{W}$ by adding the Conjugate Neuron}  \\
\begin{tabular}[c]{@{}l@{}}1. Create the $\mathcal{W}$ net from the trained $\mathcal{N}$ net. The $\mathcal{W}$ net keeps all \\neurons of the $\mathcal{N}$ net and initially has all parameters as in the $\mathcal{N}$ net.\end{tabular}                                     \\
2. Add a new conjugate neuron $\eta^{c}$ to the output layer $L$ of the $\mathcal{W}$ net.        \\
3. Link the neurons from layer $(L-1)$ to the conjugate neuron, $\eta^{c}$.\\
\begin{tabular}[c]{@{}l@{}}4. Initialize the new weights $\textbf{w}^c$  from the neurons on layer (L-1) \\ to the new conjugate output neuron $\eta^c$ and its bias $b^c$ to 0.\end{tabular}                                     \\

\textbf{Fine-Training Phase: Train the WisdomNet $\mathcal{W}$ neural network}                                                    \\
\begin{tabular}[c]{@{}l@{}}1. Use $S^r$ as the training data having label as `reject'.\end{tabular}  \\
\begin{tabular}[c]{@{}l@{}}2. Train $\mathcal{W}$ with $S^r$.  \end{tabular}                                                                          \\

\hline
\end{tabular}
}

\end{table}

{\em Step 1. Pre-train a base neural network architecture.}
Consider a neural network $\mathcal{N}$ with $(L+1)$ layers for a binary classification problem. Let $l \in \{0, \cdots, L\}$ indicate a layer of the network and $n_{l}$ denote the number of neurons at the $l^{th}$  layer. A typical neural network $\mathcal{N}$ can be expressed as
\begin{equation}
 \mathcal{N} = \mathcal{N}(n_{0}, \cdots, n_{l}, \cdots, n_{L-1}, 2),
\label{eq:typical-network}
\end{equation}

\noindent
where $n_{0}$ is the number of neurons in the input layer, and $n_{L-1}$ is the number of neurons in the last hidden layer. In this network, we have a neuron for each class label in the output layer $L^{th}$. Fig. \ref{fig-WisdomNetmodel}(a) displays a typical fully-connected neural network architecture for a binary classification problem. 
We first train a standard neural network $\mathcal{N}$ on the training data set $S$. 

{\em Step 2. Identify misclassified samples.} We generate output predictions for the training set $S$ using the trained $\mathcal{N}$ model. We identify the misclassified samples by $\mathcal{N}$ and put them into the reject set,  $S^r$. All samples in $S^r$ are re-labeled with a new `reject' label for training the WisdomNet since these misclassified samples are used to determine what the network cannot learn.

{\em Step 3. Build the WisdomNet by adding a conjugate neuron.}
The WisdomNet $\mathcal{W}$ is created from the trained base network $\mathcal{N}$ with the same number of layers (as provided in Equation (\ref{eq:typical-network})). 
However, the number of neurons in the output layer of the WisdomNet increases from $2$ to $3$ when adding the conjugate neuron (Fig. \ref{fig-WisdomNetmodel}(b)). This neuron is added in a way that the neurons in the layer before the output layer are connected to this new conjugate neuron (red arrows in Fig. \ref{fig-WisdomNetmodel}(b)). Note that the WisdomNet is built from a trained network, not from an untrained network, and the WisdomNet initially keeps all parameters of the original network. The weights connecting to the conjugate neuron from the last hidden layer are initialized to 0. The bias value of the conjugate neuron is set to 0 before the fine training phase.
The WisdomNet uses the extra conjugate neuron to recognize data instances that were incorrectly classified by the typical networks. Hence, the only difference in the architecture of the original net and the WisdomNet is the output layer (Equation (\ref{eq:WisdomNet})),

\begin{equation}
 \mathcal{W} = \mathcal{W}(n_0, n_{1}, \cdots, n_{l}, \cdots, n_{L-1}, 3).
\label{eq:WisdomNet}
\end{equation}

{\em Step 4. Fine-train the WisdomNet with only misclassified training samples.}
 We train the $\mathcal{W}$ network only on the data samples in $S^r$ that were misclassified from the training set $S$ by the $\mathcal{N}$ net. Assume that the 2 output neurons of the $\mathcal{N}$ recognize the `0' and `1' binary classes. If we label all misclassified training data as `r', then the conjugate neuron in the $\mathcal{W}$ net is trained to recognize this new label. The $\mathcal{W}$ network is initially set to keep all properties of the $\mathcal{N}$ net. After training the $\mathcal{W}$ network, it is very likely that all the misclassified data will be recognized by the conjugate neuron. It is important to note that if the $\mathcal{W}$ network rejects just the misclassified samples by the $\mathcal{N}$ network, that would be a case of overfitting. We hypothesize that there are `lucky' correct predictions by the $\mathcal{N}$ network, and they could be recognized as well.

The trained WisdomNet is used as the final classification model. 
The steps of training a WisdomNet neural network is summarized in Algorithm \ref{algo-WisdomNet}. For keeping the generality, the algorithm does not list the detail of how batches of data are used in each training epoch. The algorithm can be summarized as follows:
\begin{itemize}
    \item Line 1 - 8: The pre-training phase to construct and train a typical neural network $\mathcal{N}$ with the training set $S$.
    \item Line 9 - 16: The identification and relabeling of misclassified samples to create the reject set  $S^r$.  
    \item Line 17 - 20: The construction of the WisdomNet neural network $\mathcal{W}$.
    \item Line 21 - 26: The fine-training phase to train the WisdomNet neural network with the reject set $S^r$.
\end{itemize}

\begin{algorithm}[t!]
	\caption{WisdomNet Training Algorithm}
	\label{algo-WisdomNet}
	\renewcommand{\algorithmicrequire}{\textbf{Input:}} 
	\renewcommand{\algorithmicensure}{\textbf{Output:}} 
	\begin{algorithmic}[1]
		\REQUIRE ~~\\
		$\mathbf{S}$: training data samples with labels $\mathbf{Y}$\\$R$
		$E_{pre}$: Number of epochs in pre-training\\
	    $E_{fine}$: Number of epochs in fine-training\\
		\ENSURE
		A WisdomNet neural network model $\mathcal{W}$.\\
		\STATE\textbf{--Pre-training: Train a Standard Neural Network $\mathcal{N}$:}
		\STATE create a $\mathcal{N}$ network with $L+1$ layers;
		\STATE randomly set weights $\textbf{w}$ and bias $b$ of neurons in $\mathcal{N}$;
		\FOR{ i $<$ $E_{pre}$}
		\STATE randomly shuffle $\mathbf{S}$;
		\STATE feed $\mathbf{S}$ to the network $\mathcal{N}$; 
		\STATE update parameters of the network by applying gradient descent using backpropagation;
		\ENDFOR
		\STATE\textbf{--Generate the Set of Reject Data $\mathbf{S^r}$:}
		\FOR{\textbf{each} data $s_{i}$ $\in$ $\mathbf{S}$}
		\STATE predict output of $s_i$ as $\hat{y}_{i}$ by $\mathcal{N}$
		\IF{ $\hat{y}_{i}$ $\neq$ $y_{i}$}
		\STATE add $s_{i}$ to $\mathbf{S^r}$;
		\STATE relabel $s_i$ as $reject$
		\ENDIF
		\ENDFOR
		
		\STATE\textbf{--Construct the WisdomNet Neural Network $\mathcal{W}$:}
		\STATE $\mathcal{W}$ $\leftarrow$ $\mathcal{N}$; \slash\slash create $\mathcal{W}$ net from the trained $\mathcal{N}$ net
		\STATE add a new conjugate neuron $\eta^{c}$ to the output layer of $\mathcal{W}$;
		\STATE set all new weights $\textbf{w}^c$ and bias $b^{c}$ of $\eta^{c}$ to 0;

		\STATE\textbf{--Fine-Training: Train the WisdomNet $\mathcal{W}$:}
		\FOR{ i $<$ $E_{fine}$}
		\STATE randomly shuffle $\mathbf{S^r}$;
		\STATE feed $\mathbf{S^r}$ to the network $\mathcal{N}$; 
		\STATE update parameters of the network by applying gradient descent using backpropagation;
		\ENDFOR
		\RETURN  $\mathcal{W}$.
	\end{algorithmic}
\end{algorithm}

\subsection{Motivating Sample Scenario}
This section presents a motivating sample scenario to demonstrate the proposed WisdomNet and the training process. In this example, we generated artificial data samples using a quadratic function $F(t,v)$ shown in Equation (\ref{eq:sample-quadratic}):

\begin{equation}
    F(t,v) = -t^{2} + 0.2t - 0.5v + 0.3.
    \label{eq:sample-quadratic}
\end{equation}

\noindent
The training set $S$ contains 1024 pairs of $t$ and $v$ features. The values of $t$ and $v$ follow random uniform distribution in the range of [-1.0, 1.0]. We did the same to create another 1024 data for the test set $Q$. Each data is labeled based on the following inequality (\ref{eq:sample-quadratic}).
\begin{equation}
    F(t,v) > 0.
    \label{eq:sample-inequality}
\end{equation}

\begin{figure}[t]
\centering
\includegraphics[width=0.23\textwidth]{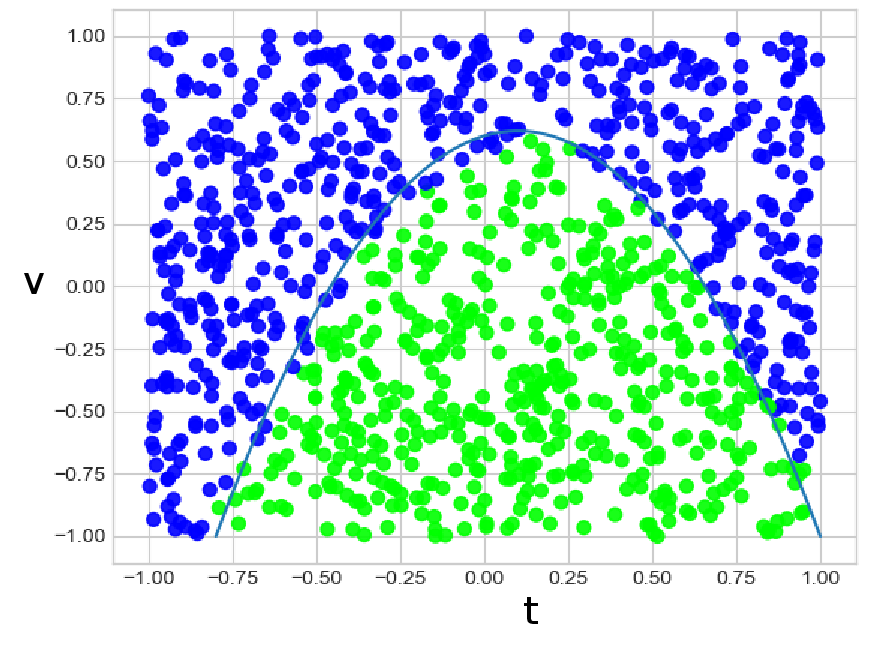}
\includegraphics[width=0.23\textwidth]{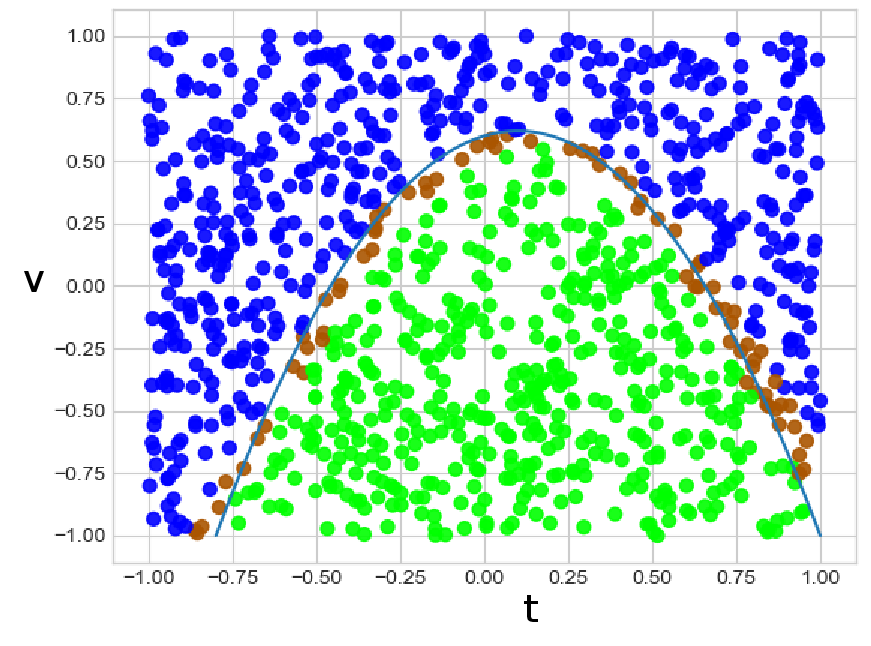}
~~~~~~~~~~~~~~~~~~~~~~~~~~~~~~~~~~~~~~(a)~~~~~~~~~~~~~~~~~~~~~~~~~~~~~~~~~~~~~~~~~~(b)
\caption{Sample prediction results on the testing data set of the standard network $\mathcal{N}$ and the WisdomNet $\mathcal{W}$. a) About $98\%$ accuracy by $\mathcal{N}$ with misclassifications visible around the parabola (some blue points are under the parabola, and some green points are above the parabola as misclassifications), b) Zero misclassification by $\mathcal{W}$ with about $5\%$ reject rate (no blue points under the parabola and no green points are above the parabola; rejected cases are marked with brown color).}
\label{fig-quad-WisdomNet}
\end{figure}

First, we trained the $\mathcal{N}(2,4,2)$ network with the $S$ data set. For the testing result, we achieved about $98\%$ accuracy of the network $\mathcal{N}$ on the $Q$ testing set. In Fig. \ref{fig-quad-WisdomNet}(a), it is possible to see misclassifications around the parabola. The instances below the parabola curve satisfy the inequality (\ref{eq:sample-inequality}) and should be colored as green, whereas the instances above the parabola should be colored as blue. 

Then, we fed the original training set $S$ to the network $\mathcal{N}$ and recorded all misclassified training data. We used these misclassified samples to create the reject set. We built the WisdomNet $\mathcal{W}(2,4,3)$ from the trained network $\mathcal{N}$ and added the conjugate neuron to the output layer. The conjugate neuron was initialized with zero weights and bias. All other neurons were kept the same as on the original network. We trained the WisdomNet $\mathcal{W}(2,4,3)$ on the reject set.   

As of the prediction results on the testing set $Q$, the WisdomNet was able to classify the data with zero error, as shown in Fig. \ref{fig-quad-WisdomNet}(b). The percentage of rejected samples by the WisdomNet is about $5\%$ (slightly greater than the $2\%$ misclassified testing samples by the network $\mathcal{N}$). None of the data was misclassified.

\subsection{Justification}
We have two observations: 1) the WisdomNet $\mathcal{W}$  must inherit all properties of the original base network $\mathcal{N}$; 2) the WisdomNet is able to converge to zero training error in the fine-training process.  
The first observation states that after adding a conjugate neuron but before training $\mathcal{W}$ on misclassified samples, the WisdomNet $\mathcal{W}$ is equivalent to the original network $\mathcal{N}$. The second observation states it is possible for the WisdomNet to learn the data that the base network misclassifies.

\hfill
\newline
\textbf{Preliminaries:} A neural network, $\mathcal{N}$, with $(L+1)$ layers is optimized on a finite set of $M$ $d_0$-dimensional training data, $\{(\textbf{x}_{i},y_{i})\}_{i=1}^M$, where the input samples are $\textbf{X}=[\textbf{x}_{1},...,\textbf{x}_{M}]\in \mathbb{R}^{d_0 \times M}$, and their corresponding outputs are  $\textbf{Y}=[y_{1},...,y_{M}]\in \mathbb{R}^{1 \times M}$. Let $n_l$ denote the the number of neurons at the $l^{th}$ layer ($n_0=d_0$). The activation function is represented with $\sigma$. The weight matrix connecting layer $(l-1)$ to layer $l$ is represented with $\textbf{W}_l\in\mathbb{R}^{n_{l-1}\times n_l}$ and the bias vector for the $l^{th}$ layer is represented with $\textbf{b}_l\in \mathbb{R}$. Then, $\textbf{W}_L\in\mathbb{R}^{n_{L-1}\times n_L}$ and $\textbf{b}_L\in \mathbb{R}$ denote the weight matrix and bias vector for the output layer respectively.
The feedforward neural network maps the input $\textbf{x}$ to category $y$ though the function of $\textbf{x}$ parameterized by $\bm{\theta}$,

\begin{equation}
    \resizebox{.95\hsize}{!}{$f(\textbf{x};\bm{\theta}) = \sigma(\textbf{W}^{\top}_{L}\sigma(\textbf{W}^{\top}_{L-1}\sigma(\cdots \sigma(\textbf{W}^{\top}_{1}\textbf{x} + \textbf{b}_{1}) + \cdots) + \textbf{b}_{L-1})+ \textbf{b}_{L})$}
\label{eq:output}
\end{equation}

\noindent
where $\bm{\theta}$ consists of weights and bias parameters. The network learns the parameters $\bm{\theta}$ that results the best mapping of $y=f(\textbf{x};\bm{\theta})$.

\hfill
\newline
\textbf{Theorem 1 (Equivalence)}: \textit{
After adding the conjugate neuron, the WisdomNet,  $\Tilde{f}(\textnormal{\textbf{x}};\tilde{\bm{\theta}})$, is equivalent to 
 the base network, $f(\textnormal{\textbf{x}};\bm{\theta)}$, i.e., $\forall \textnormal{\textbf{x}} \in\mathbb{D}$,  $\Tilde{f}(\textnormal{\textbf{x}};\tilde{\bm{\theta}})=f(\textnormal{\textbf{x}};\bm{\theta)}$, where $\mathbb{D}$ is the domain of the input set.}

The proof of theorem can be done in two steps. Firstly, we show that there is no change in the outputs of the neurons inherited from the base network.

The WisdomNet, $\Tilde{f}(\textbf{x};\tilde{\bm{\theta}})$, adds a new conjugate neuron, $\eta^c$ to the output layer, thus adding a weight vector, $\textbf{w}^c$, coming from the $(L-1)^{th}$ layer and a new bias scalar $b^c$. At the initialization, the only differences in parameter set $\Tilde{\bm{\theta}}$ compared to $\bm{\theta}$ are the weight matrix $\Tilde{\textbf{W}}_L$ and the bias vector $\Tilde{\textbf{b}}_L$ coming into the output layer:

\begin{equation}
    \Tilde{\textbf{W}}_L = \left[ \begin{array}{c} \textbf{W}_L |  \textbf{w}^c\end{array}\right]
\label{eq:weight}
\end{equation}

\begin{equation}
    \Tilde{\textbf{b}}_L = \left[ \begin{array}{c} \textbf{b}_L \\ b^c\end{array}\right].
\label{eq:bais}
\end{equation}

The initialized WisdomNet, $\Tilde{f}(\textbf{x};\tilde{\bm{\theta}})$, can be expressed as follows:
\begin{equation}
   \resizebox{.95\hsize}{!}{$\Tilde{f}(\textbf{x};\tilde{\bm{\theta}}) =  \sigma(\Tilde{\textbf{W}}^{\top}_{L}\sigma(\textbf{W}^{\top}_{L-1}\sigma(\cdots \sigma(\textbf{W}^{\top}_{1}\textbf{x} + \textbf{b}_{1}) + \cdots) + \textbf{b}_{L-1})+ \Tilde{\textbf{b}}_{L})$ }
\label{eq:wisdom}
\end{equation}

There are no changes in the intermediate layers up to the last layer, hence the activations $\textbf{a}_{L-1}$ of layer $(L-1)$ are unchanged. On the last layer $L$, the outputs of neurons inherited from the base network are based on the original weights $\textbf{W}_L$ and biases $\textbf{b}_L$ and the activations $\textbf{a}_{L-1}$. Thus these outputs are the same for the WisdomNet and the base network.  
\begin{equation}
    \textbf{a}_L=\sigma(\textbf{W}_L\textbf{a}_{L-1}+\textbf{b}_L)
\end{equation}

Secondly, we show that the conjugate neuron $\eta^c$ is not activated. Since $\textbf{w}^c$ and $b^c$ are set to 0, the output of $\eta^c$ in the output layer, $a^c_L$, is also 0.

\begin{equation}
\begin{array}{l}
    \textbf{w}^c = \textbf{Zeros}(n_{L-1},1),\\
    b^c=0
\end{array}
\label{eq:conjugate-weight}
\end{equation}

\begin{equation}
    a^c_L=\sigma(\textbf{w}^c\textbf{a}_{L-1}+b^c)=0
\end{equation}

This theorem shows that the $\mathcal{W}$ net inherits all the knowledge from the $\mathcal{N}$ net by maintaining its parameters and initially keeping the conjugate neuron inactive.

\hfill
\newline
\textbf{Theorem 2 (Convergence)}: \textit{The WisdomNet $\mathcal{W}$ is able to converge to zero training error for reject samples on differentiable local minima if $|S^r|<n_{L-1}n_{L-2}$, where $|S^r|$ is the number of reject samples, since i) $|S^r| \ll M$ 
and ii) the number of reject samples $|S^r|$ is expected to be lower than the product of $n_{L-1}$ and $n_{L-2}$}.

After pre-training, the number of misclassified training samples will be relatively few for a highly accurate neural network. These misclassified training samples are used to create the new training set $S^r$ for the new WisdomNet network. Hence, the size of $S^r$ is much smaller than the size $M$ of the original training set. Thus, the new neural network can be considered as over-parameterized for the set $S^r$, and for over-parameterized, it is possible to achieve zero training error \cite{allen2018convergence}. 
Our theorem is also based on the proof by Soudry and Carmon that can be stated in our case as if $|S^r|\leq n_{L-1}n_{L-2}$, then all differentiable local minima are also global minimal with zero training error \cite{soudry2016no}.  

In the fine-training, the misclassified samples from the pre-training set are put into the set of reject, $S^r$. The set $S^r$ is a small subset of the original training data set. Since the labels of data in $S^r$ are updated to `reject', the new cost function can be represented as

\begin{equation}
C = \frac{1}{2} \sum_x \left\|\Tilde{y}(\textbf{x})-\Tilde{f}(\textbf{x};\Tilde{\theta})\right\|^2 = 0,
\label{eq-cost}
\end{equation}

\noindent
where $\Tilde{y}(\textbf{x})$ only holds a single label for $\textbf{x} \in S^r$ as reject.

Assume that the activations of the original neurons are represented with $a^0$ and $a^1$, and the new conjugate neuron activation is denoted with $a^c$. 
The partial derivatives vector for the total error cost of the $\mathcal{W}$ network at the output layer can be represented as 

\[
\nabla_a C = \Big[\frac{\partial{}C}{\partial{}a^0_L},\frac{\partial{}C}{\partial{}a^1_L}, \frac{\partial{}C}{\partial{}a^c_L} \Big], 
\]

\noindent
where the $\nabla_a C$ vector expresses how the cost $C$ will change concerning the output activation values.

Initially, the $\mathcal{W}$ network is equivalent to the $\mathcal{N}$ network, as provided by Theorem 1. The error at the output nodes is computed by the product of the partial derivative of the cost by the activation and the derivative of the activation function. Since the derivative of an activation function is not negative, the sign of the error depends on the partial derivative.
When feeding the misclassified training data to the $\mathcal{W}$ network, either one of the original output neurons will activate, and the conjugate neuron $\eta^c$ will be inactive. Therefore, in the backpropagation, the gradient of the error cost with respect to the activation $a^c$ of the conjugate neuron will have an opposite direction compared to the other with respect to the original output neurons activations, $a^0$ and $a^1$. This direction difference will be consistent for every sample fed to the WisdomNet network for the first time. In this fine-training, the direction of the gradient will not oscillate as in training traditional networks with varying class labels.

As provided in Theorem 2, the WisdomNet $\mathcal{W}$ is able to converge to zero training error in the fine-training process. Assuming that the training set $S$ and the testing set $Q$ come from the same distribution, then the $\mathcal{W}$ network is able to use the conjugate neuron not to misclassify any data in $Q$ that were incorrectly classified by the base network $N$. We should note that during the fine-training process all parameters of the network are updated. Hence, some of the samples which were correctly classified now can be classified as reject. This is a healthy outcome that shows that most likely the fine-training process does not overfit the misclassified data. Moreover, our system is able to determine lucky predictions made by the network which could lead to incorrect classifications in real scenarios. 

\section{Experiments and Results} \label{experiment}

In this section, we evaluate the performance of the WisdomNet on diverse data sets using a variety of neural network architectures to show that the WisdomNet is generalizable and able to yield zero error in different contexts. We conducted experiments using our code implemented in Python 3.

We have studied the following network architectures to show the generalizability of the WisdomNet:
\begin{itemize}
\item Basic multilayer perceptron (MLP) with one hidden layer,
\item Convolutional neural network (ConvNet), and
\item Deep network architecture such as ResNet.
\end{itemize}

We have experimented the WisdomNet on the following diverse data sets:
\begin{itemize}
\item Synthetic data: Artificially generated data including MOON, CIRCLE, Gaussian Quantile, and XOR,
\item Standard classification data: Banknote Authentication (BN), Ionosphere (IO), and Breast Cancer Wisconsin (BC) data set from UCI \cite{Uci},
\item MNIST: A benchmark data set of handwritten digits \cite{lecun1998gradientmnist}, and
\item Dogs vs Cats: A color images set of dogs and cats from Kaggle \cite{Kaggle}. 
\end{itemize}

After providing evaluation measures in the following section, we explain how the WisdomNet performs on various network architectures using the corresponding data sets. We also present the study on the factors that influence the performance of the WisdomNet.

\subsection{Evaluation Measures}
\begin{figure}[t]
\centerline{\includegraphics[width=0.4\textwidth]{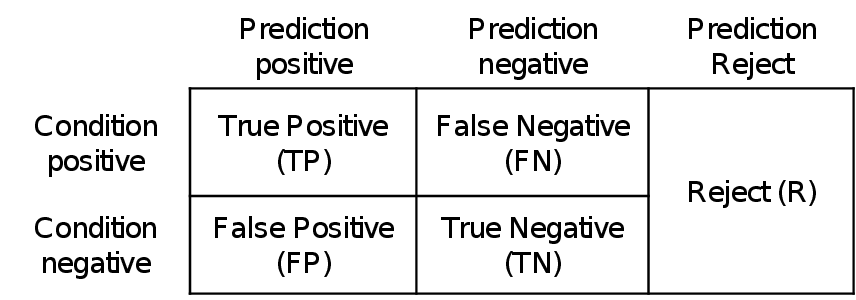}}

\caption{The confusion matrix of the WisdomNet with prediction reject column}

\label{fig-confusion-matrix}
\end{figure}

We are going to provide our results using traditional evaluation measures such as accuracy since it is the most commonly used evaluation measure \cite{fatourechi2008comparison,sigdel2013pacc}. Since our interpretation of trustable learning introduces a new category of predicted results as reject, we will update a few of the measures to integrate `reject' samples. We denote $R$ as the number of samples classified as reject. The total number of samples is $N=TP+TN+FP+FN+R$, where $TP$, $FP$, $FN$, and $TN$ are the number of the true positive, false positive, false negative, and true negative samples, respectively. The confusion matrix of the WisdomNet includes a reject column, as shown in Fig. \ref{fig-confusion-matrix}. 

Moreover, we also used the total error rate (\ref{eq:acccuracy5}) and reject rate ($R_{rate}$) (\ref{eq:reject-rate}) to evaluate the network performance: 

\begin{equation}
\label{eq:acccuracy5}
Error=\frac{FP + FN}{N}
\end{equation}

\begin{equation}
\label{eq:reject-rate}
R_{rate}=\frac{R}{N}
\end{equation}

\subsection{WisdomNet on Basic Multilayer Perceptron Neural Network}
We first tested our WisdomNet on the basic MLP network architecture with one hidden layer. We used the BN, IO, and BC standard data sets to compare the network performances. For each data set, we held out some data for training and used the rest for testing, as provided in Table \ref{tab:MLP_data}. 

In each neural network, the number of input neurons is equal to the number of features in each data set. There are two output neurons in each MLP network since these problems are binary classification problems. In the hidden layer, there are 16 neurons for the BC and IO data sets and 4 neurons for the BN data set (Table \ref{tab:MLP_data}). After training the MLP models, the WisdomNet architecture increases the number of output neurons by adding one more conjugate neuron. Hence, there are three output neurons in the WisdomNet networks. The weights coming to the conjugate neuron and its bias are set to 0 at first. Algorithm \ref{algo-WisdomNet} is used to train the WisdomNet.

\begin{table}[t]
\begin{center}
\caption{ A summary of the standard binary classification  data sets and networks architectures.}
\label{tab:MLP_data}
\begin{tabular}{l|ccccc}
\hline
\hline
Data sets & Attributes & Train & Test & MLP & WisdomNets\\ \hline
BC & 30  & 512  & 57 & 30-16-2 & 30-16-3  \\
BN & 4  & 1,235  & 137 & 4-4-2 & 4-4-3  \\
IO & 32  & 270  & 81 & 32-16-2 & 32-16-3\\
 \hline
\end{tabular}
\end{center}

\end{table}

Table \ref{tab:MLP_Wisdom} shows the performance of the MLP and the WisdomNet for the data sets mentioned in Table \ref{tab:MLP_data}. The middle four columns show the classification error rates and the corresponding accuracies of the MLP networks and the WisdomNet. The last column presents the reject rate of the WisdomNet when classifying the test data. The MLP neural networks can classify all testing data with a high accuracy rate of above $96\%$. The highest accuracy of the MLP is about $99\%$ when classifying the BN data. However, all these MLP networks have a classification error between around $1\%$ and $3.71\%$.      
As shown in the table, the WisdomNet reaches $0\%$ for the classification error rate in all tested data sets. It successfully moved all samples hard to classify to `reject', but some correctly classified samples are also labeled as reject. The reject rates for the IO and BC data sets are about $13.58\%$ and $10.05\%$,  respectively. The lowest reject rate is just around $1.5\%$ for the BN  data set. 

\begin{table}[t]
  \centering
  \caption{The testing performance of the WisdomNet vs MLP on the standard binary classification  data.}
      \vspace{-5pt}
    \begin{tabular}{c|cc|ccc}
\multicolumn{1}{c}{} & \multicolumn{2}{c}{MLP} & \multicolumn{3}{c}{WisdomNets} \\
   \cmidrule(lr){2-3}\cmidrule(lr){4-6}
    Data sets & Error & Accuracy & Error& Accuracy & $R_{rate}$  \\ \hline
    BC & 1.76\% & 98.24\% & 0\% & 100\%    & 10.05\% \\
    BN & 1\% & 99\% & 0\% & 100\%    & 1.5\% \\
    IO & 3.71\% & 96.29\% & 0\% & 100\%    & 13.58\% \\\textbf{}

    \end{tabular}%
    \vspace{-10pt}
  \label{tab:MLP_Wisdom}%
\end{table}%

\subsection{WisdomNet on Convolutional Neural Networks }\label{section-cnn}
\begin{figure}[t]
\centerline{\includegraphics[width=0.45\textwidth]{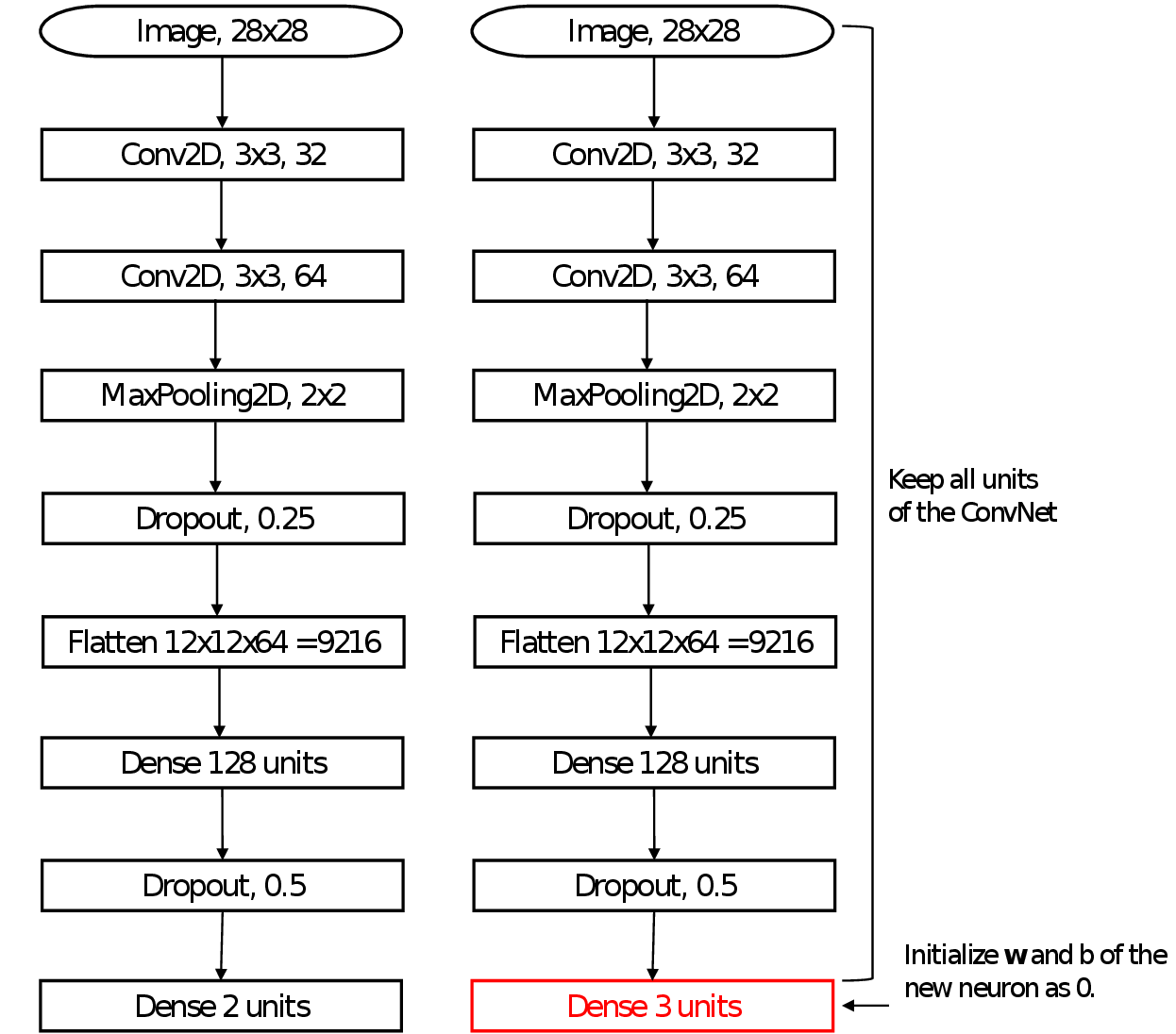}}

\caption{WisdomNet architecture based on deep ConvNet. \textbf{Left}: The ConvNet network. \textbf{Right}: The WisdomNet: the red output layer contains a conjugate neuron having weights and bias initialized as 0; all parameters of the trained ConvNet model are initially transferred to the WisdomNet.}

\label{fig-cnn}
\end{figure}

In this part, we applied the WisdomNet on the deep convolutional neural network architecture shown in Fig. \ref{fig-cnn}. The ConvNet is composed of feature layers and classification layers. For the feature groups, we used 2 convolution layers followed by a max-pooling and a flattening layer. To avoid overfitting, we used dropout with $p = 0.25$ between the max-pooling and the flattening layers. For the classification, we used two dense layers (128-2) with a dropout of 0.5.

The WisdomNet is constructed from the trained ConvNet base network. As laid out in the Algorithm \ref{algo-WisdomNet},  a conjugate neuron is added to the network. Initially, the weights and bias of the conjugate neuron are set to 0 while keeping all parameters of the base network. 

\begin{figure}[t]
\centerline{\includegraphics[width=0.4\textwidth]{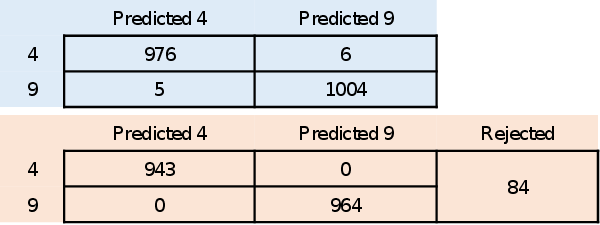}}

\caption{Confusion Matrix of ConvNet (Top) vs WisdomNet (Bottom) on MNIST 4-9 sub-set}

\label{fig-cnn-confusion}
\end{figure}

We worked on a binary classification problem where misclassifications are likely to occur. We picked the digits 4 and 9 from the MNIST data to create a binary data set. In this experiment, the training set has  11791 images of the digits 4 and 9. The test set contains 1991 images. 
The base ConvNet model was able to reach around 99.4\% accuracy with about $0.6\%$ classification error rate. Applying the WisdomNet reduced the error to $0\%$ with about $4.2\%$ reject rate. 
Fig. \ref{fig-cnn-confusion} shows the confusion matrix of the two networks. 
Despite high true positive and true negative results, the convolutional net made misclassifications on 11 instances. 
The WisdomNet returned 84 digits as hard to classify to obtain 0 false positives and 0 false negatives.    

\subsection{WisdomNet on Deeper Networks and Transfer Learning}
We also applied the WisdomNet on a deeper network than the networks in previous experiments. Many deep learning models are available such as VGG16, VGG19 \cite{simonyan2014VGG}, ResNet34, and ResNet50 \cite{he2016residualdeep}. We have picked the ResNet50 (a 50-layer deep Residual Network), as the base model as an example of a deep learning network since it can be considered as fairly deep compared to other architectures. The ResNet50 was trained on millions of images of the ImageNet \cite{imagenet_cvpr09} data sets to classify 1000 object categories.  

In these experiments, we used a new data set of cats and dogs from Kaggle \cite{Kaggle}. The original data set contained 25,000 images, but we used just 4000 images (first 2000 cat and first 2000 dog images) for training and 400 images for validation (200 from each class). This is a challenging deep learning problem since the training data set is small. We applied transfer learning and fine-tuning to solve this problem as suggested popularly (such as Keras\cite{chollet2015keras}).      

For evaluation, we will perform two comparisons. Firstly, we are going to apply transfer learning using the ResNet50 to classify the images of dogs and cats. Normally, the ResNet50 should already be able to classify dogs, cats, and other class labels based on the training using ImageNet. In this case, we truncate the last layer of the ResNet50 and use the remaining layers to create the bottleneck features of the data. After removing the last layer of the ResNet50, we run this model on the data set of cats and dogs (both training and validation) once and then maintain the feature sets provided by the model. Since our task is a binary classification of the dogs-cats problem, we use a new top network to classify the 2 categories. The new top network is a classifier with multiple fully-connected layers. Finally, we run backpropagation on the fully-connected layers using the extracted bottleneck features.        

Secondly, we apply the fine-tuning method to see if it helps to classify dogs and cats. In this case, we re-train the last residual block of the ResNet50 and the classifier layers. All other predecessor layers in the residual network were frozen by not updating their weights during the fine-tuning process. We expect the network to improve its performance by updating parameters from the top network to the last residual block.
 
\begin{figure}[t]
\centerline{\includegraphics[width=0.5\textwidth]{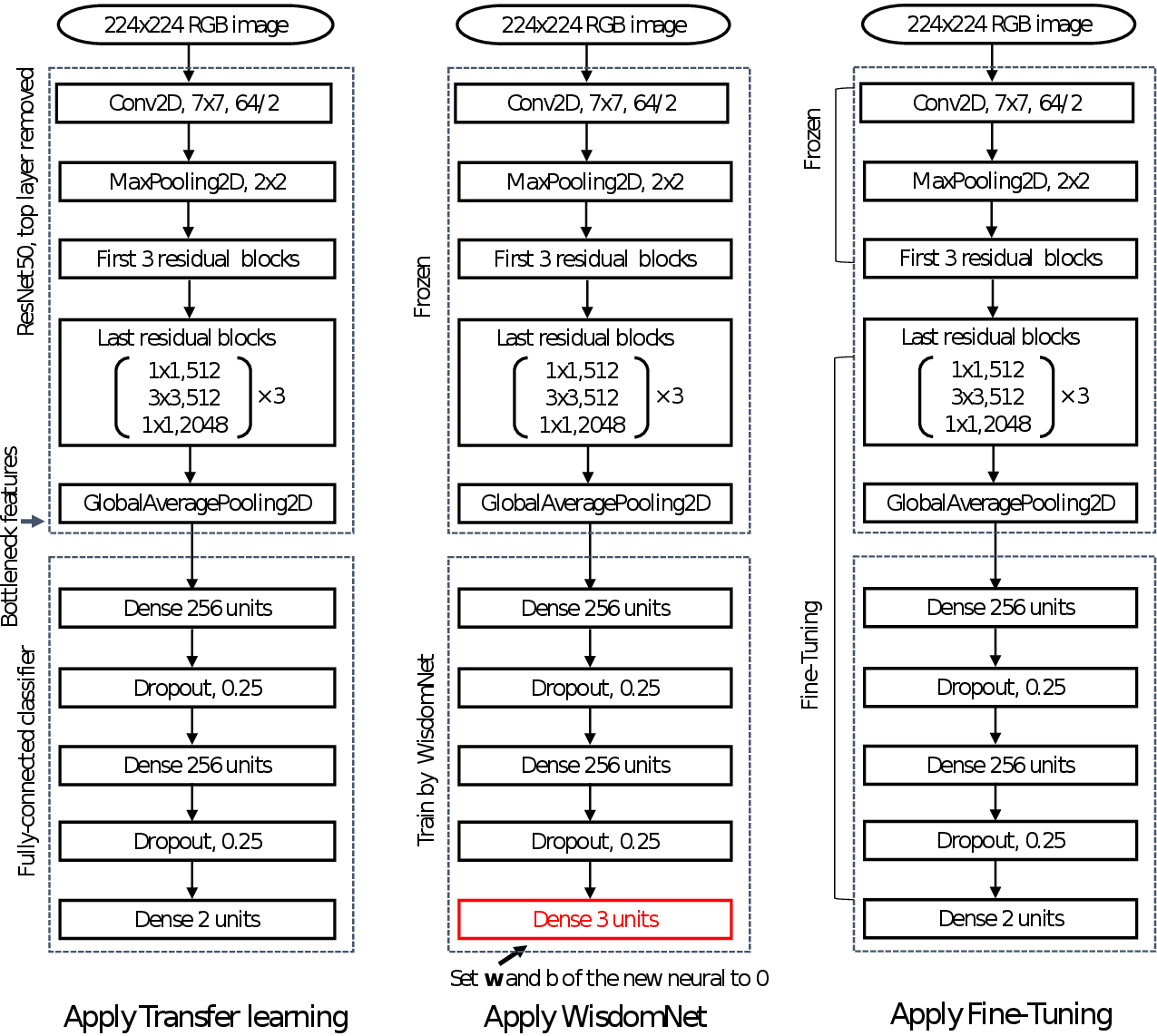}}

\caption{Network architectures using 50-layer residual as the base model. \textbf{Left}: The transfer learning model. \textbf{Middle}: The WisdomNet model: the red output layer contains a conjugate neuron with 0 initialized weights and bias; all weights and bias of other units are initially inherited from the transfer learning model. \textbf{Right}: The fine-tuned model.}

\label{fig-deep_resnet50}
\end{figure}

We build the WisdomNet using the transferred learning model of the ResNet50  with two output classes as the base network. 
We add a conjugate neuron to the output layer. All misclassified training data items by the base network were labeled as reject and used to train the WisdomNet.    
Figure \ref{fig-deep_resnet50} shows the three network architectures. We used Keras deep learning library with the TensorFlow in the backend in this experiment. We trained and validated the models on a system with GPU GeForce GTX 1080 Ti with compute capability 6.1. Each training image took less than 1ms to process by the WisdomNet. The fine-tuned network used around 100ms for processing a single image.

\begin{figure}[t]
\centerline{\includegraphics[width=0.47\textwidth, trim=8mm 8mm 0mm 0mm]{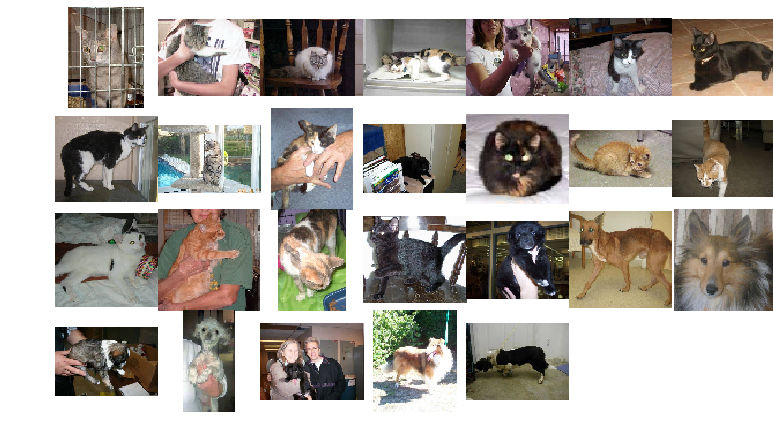}}

\caption{Images of dogs and cats that the WisdomNet rejected to make a decision.}
\vspace{-10pt}
\label{fig-u-cat-dog}
\end{figure}

Table \ref{tab:deep-res-result} shows the performance of the three models. The WisdomNet outperforms other models in reducing the misclassification error rate on the validation data. The deep transfer learning network has an error rate of $2.5\%$, whereas the fine-tuned network has an error rate of around $2\%$. On the other hand, the WisdomNet does not misclassify any data. The percentage of `reject' samples is just $6.5\%$. 
A set of images that the WisdomNet rejected to make a decision is provided in Fig.\ref{fig-u-cat-dog}. 

\begin{table}[t]
\begin{center}
\caption{The testing results of the WisdomNet vs deep learning models on the dogs-cats images classification problem.}
\label{tab:deep-res-result}

\begin{tabular}{l|cc}
\hline
\hline
Model & Error & Reject\\ \hline
ResNet50 after transfer learning & 2.5\%  & -\\
\makecell[c]{Applying WisdomNet on the transferred\\ learning model of ResNet50} & 0  & 6.5\%\\
\makecell[c]{Fine-tuning the ResNet50 after transfer\\ learning model} & 2\%  & -\\
 \hline
\end{tabular}

\end{center}

\end{table}

\subsection{Factors Affecting Trustable Learning}
In this section, we study the factors that affect the performance of the WisdomNet: the accuracy of the base network, the similarity of training data, and the noisy data.

\subsubsection{Base Network Performance} \hfill\\
The WisdomNet is constructed on top of other neural networks. In this section, we inspect how the performance of the base network affects the performance of the WisdomNet. 

By training a neural network on the same data set but changing the number of epochs allows us to get different base network performances. We generated two pairs of training and testing sets using the hold-out method on the BN data:
\begin{itemize}
\item BN80: The first $80\%$ of the instances were used for training, and the remaining 20\% were used for testing.

\item BN90: The first $90\%$ of the instances were used for training, and the remaining 10\% were used for testing.
\end{itemize}

We used a three-layer MLP network $\mathcal{N}(4,4,2)$ on both BN80 and BN90 data. Then, we evaluate the performance of the WisdomNet $\mathcal{W}(4,4,3)$.

\begin{figure}[t]
\centerline{\includegraphics[width=0.5\textwidth]{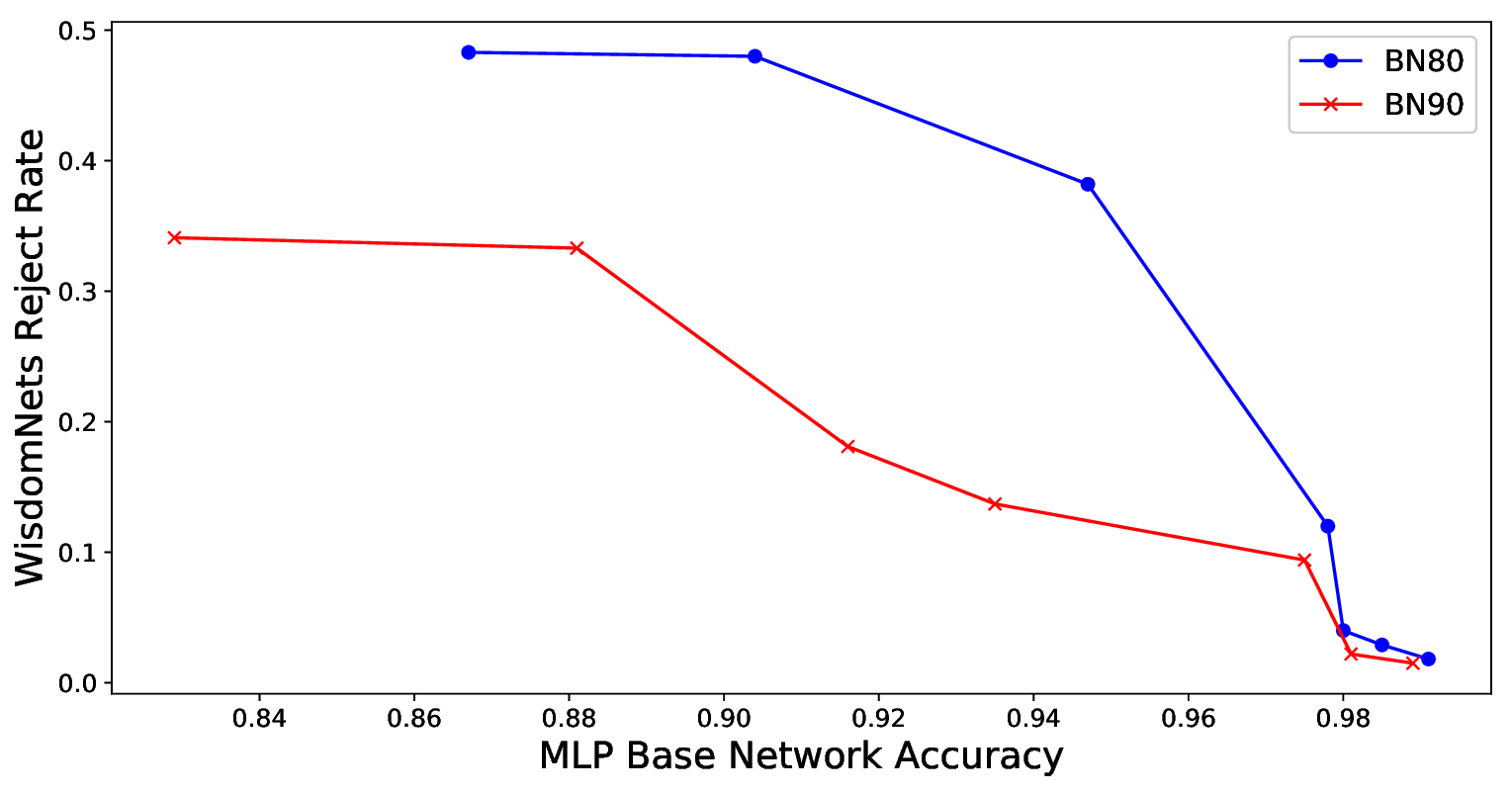}}

\caption{Effect of base network accuracy}

\label{fig-BN_Plot}
\end{figure}

Fig. \ref{fig-BN_Plot} shows the experiment results. The WisdomNet used the MLP networks with varying accuracy from around $82\%$ to above $99\%$. It can be observed that if an MLP network with accuracy less than $91\%$ on the BN80 data is used as the base network, while the WisdomNet may reduce the error to 0, the reject rate goes up to $50\%$. Such a high reject rate is not useful at all. The WisdomNet produces many `reject' data samples when the base network accuracy goes below $90\%$. When the MLP network had lower than $89\%$ accuracy on BN90 data, the WisdomNet rejected more than $30\%$ of the test samples. 

As the performance of the MLP network is increased, the reject rate of the WisdomNet decreases. The WisdomNet only rejects less than $10\%$ of testing data instances when the base network has an accuracy of around 98\%. For a higher accurate MLP base network, the WisdomNet is able to reject only around $1\%$ of the testing samples. The WisdomNet expects a fairly accurate base network to maintain the reject rate within $10\%$.   

In these experiments, the reject rate is larger than the rate of misclassified samples of the base network. This shows that the WisdomNet is not overfitting the reject set, and meanwhile, it does not necessarily enforce a prediction for every sample as the base network.
Moreover, since some of the correctly classified samples by the base network are rejected, the WisdomNet may potentially avoid lucky predictions made by the base network. 

There are two important issues. Firstly, if the rate of misclassified samples is high, then the network may start to reject most of the data. Secondly, if there are very few or zero misclassified samples and these samples may not actually be representative of what the network cannot actually classify, then the network may not realize those probable misclassifications in test data.

\subsubsection{Similar Samples of Different Classes}\hfill\\
When similar samples belong to different categories, it becomes harder for a neural network to classify them correctly. 
In this experiment, we examine how the similarity of samples belonging to different classes affect the WisdomNet given the base network has fairly high accuracy. Even though the accuracies of base networks are similar, the reject rates might differ. 
We created different sets of handwritten digits from the MNIST data. 
Here we used 10 different pairs of digits to create our sub-datasets: MNIST(0-1),  MNIST(1-7),  MNIST(1-9), MNIST(2-3),  MNIST(2-7),  MNIST(3-5),  MNIST(4-9),  MNIST(6-9),  MNIST(7-8),  MNIST(8-9). While some selected pairs may contain data easy to classify (e.g., 0-1), they also have a few confusing cases. There are some pairs of digits with similarity shape (e.g., 2-3) that may harder to distinguish.   
The distribution of data samples for each sub-dataset is listed in Table \ref{tab:MNIST_subsets}.

\begin{table}[b]
\vspace{-10pt}
\begin{center}
\caption{ A summary of the subset of MNIST. }

\label{tab:MNIST_subsets}
\begin{scriptsize}
\begin{tabular}{l|cccc}
\hline
\hline
Data sets & Digits & Input size & Train Data & Test Data \\ \hline
MNIST(0-1) & 0 and 1 & 28 x 28  & 12665  & 2,115  \\
MNIST(1-7) & 1 and 7 & 28 x 28  & 13007  & 2,163  \\
MNIST(1-9) & 1 and 9 & 28 x 28  & 12,691  & 2,144  \\
MNIST(2-3) & 2 and 3 & 28 x 28  & 12,089  & 2,042  \\
MNIST(2-7) & 2 and 7 & 28 x 28  & 12,223  & 2,060  \\
MNIST(3-5) & 3 and 5 & 28 x 28  & 11,552  & 1,902  \\
MNIST(4-9) & 4 and 9 & 28 x 28  & 11,791  & 1,991  \\
MNIST(6-9) & 6 and 9 & 28 x 28  & 11,867  & 1,967  \\
MNIST(7-8) & 7 and 8 & 28 x 28  & 12,116  & 2,002  \\
MNIST(8-9) & 8 and 9 & 28 x 28  & 11,800  & 1,983  \\
 \hline
\end{tabular}
\end{scriptsize}
\end{center}
\end{table}

\begin{figure}[t]
\centerline{\includegraphics[width=0.5\textwidth]{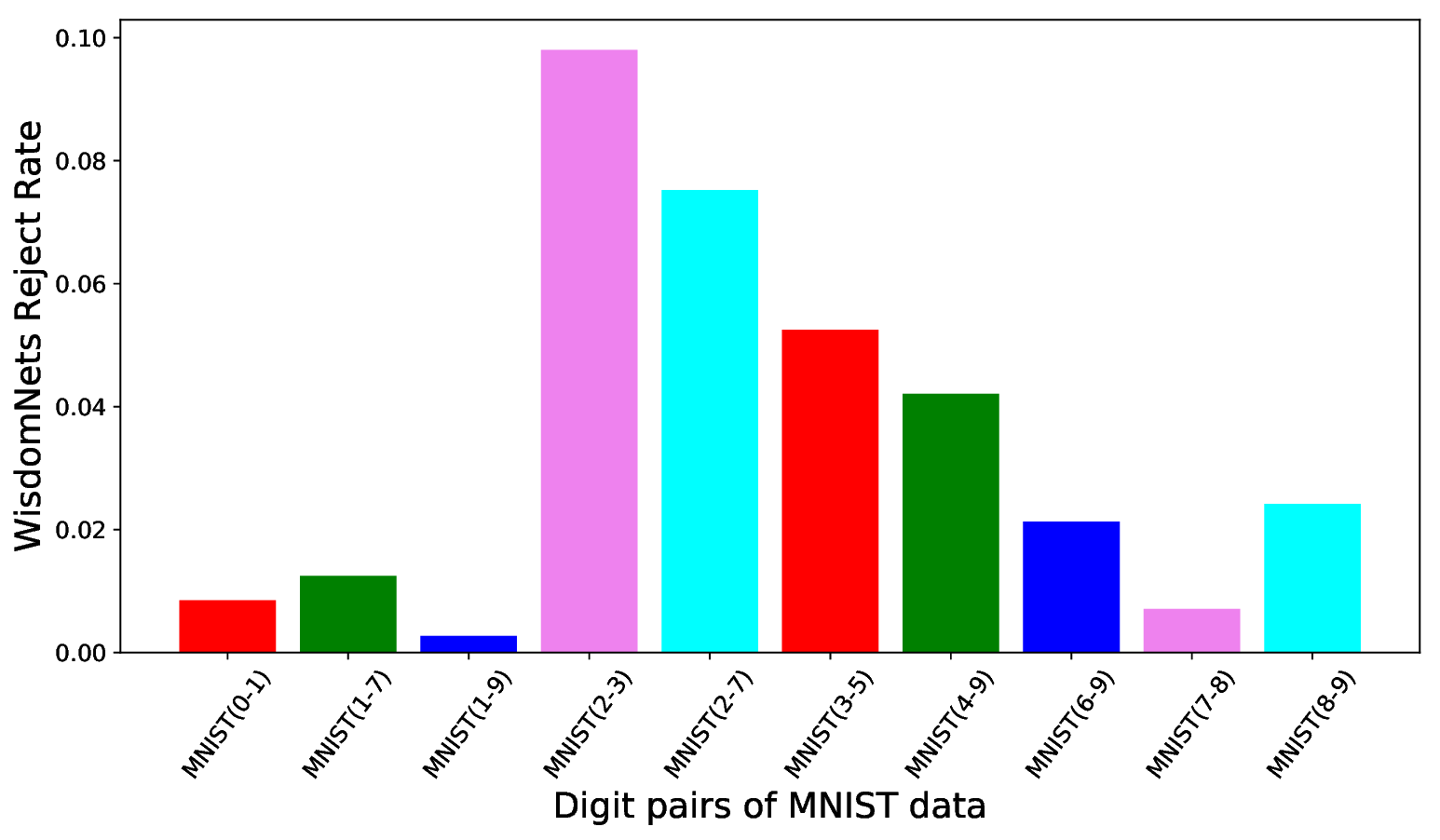}}
\caption{Effect of data set}

\label{fig-MNIST-Bar}
\end{figure}

We used the same architectures of the WisdomNet and ConvNet network as shown in Fig. \ref{fig-cnn} in Section \ref{section-cnn}. For each data set, the ConvNet model was trained to acquire above $99\%$ accuracy. 
Then, the WisdomNet was trained based on the trained ConvNet to have a zero prediction error on each data set separately while labeling some data as reject. 
The rate of reject samples is shown in Fig. \ref{fig-MNIST-Bar}. On average, the WisdomNet rejects less than 4\% of the test data. For the data sets MNIST(2-3), MNIST(2-7), MNIST(3-5), and MNIST(4-9), the reject percentage is above 4\%. The MNIST(2-3) data set has the most rejected samples with less than $10\%$. 
The reject percentages for MNIST(6-9) and MNIST(8-9) are around $2\%$. 
There is a low reject rate for the data samples in MNIST(1-7)(around $1\%$), MNIST(0-1)(under $1\%$), MNIST (7-8) (under $1\%$), and MNIST(1-9)(under $0.5\%$).  

\subsubsection{Noisy Data}\hfill\\
In many applications, the noise can be represented with Gaussian distribution, although some applications can have more complex distributions. 
In this section, we explore the effect of adding Gaussian noise to the training data on the performance of the WisdomNet. 
The Gaussian distribution is specified by two factors, the mean $\mu$ and the standard deviation $\sigma$. 
In the normal distribution, approximately $68.27\%$, $95.45\%$, and $99.73\%$ of the data are within $\pm1\sigma$, $\pm2\sigma$, and $\pm3\sigma$ of the mean, respectively.

We used the MOON, CIRCLE, and XOR synthetic data of varying shapes. The base MLP network for the CIRCLE data is N(2,4,2), whereas  N(2,8,2) networks are used for the MOON and the XOR data. The WisdomNets are created and trained based on the trained MLP networks.   

\begin{figure*}[ht]
\centerline{\includegraphics[width=\textwidth]{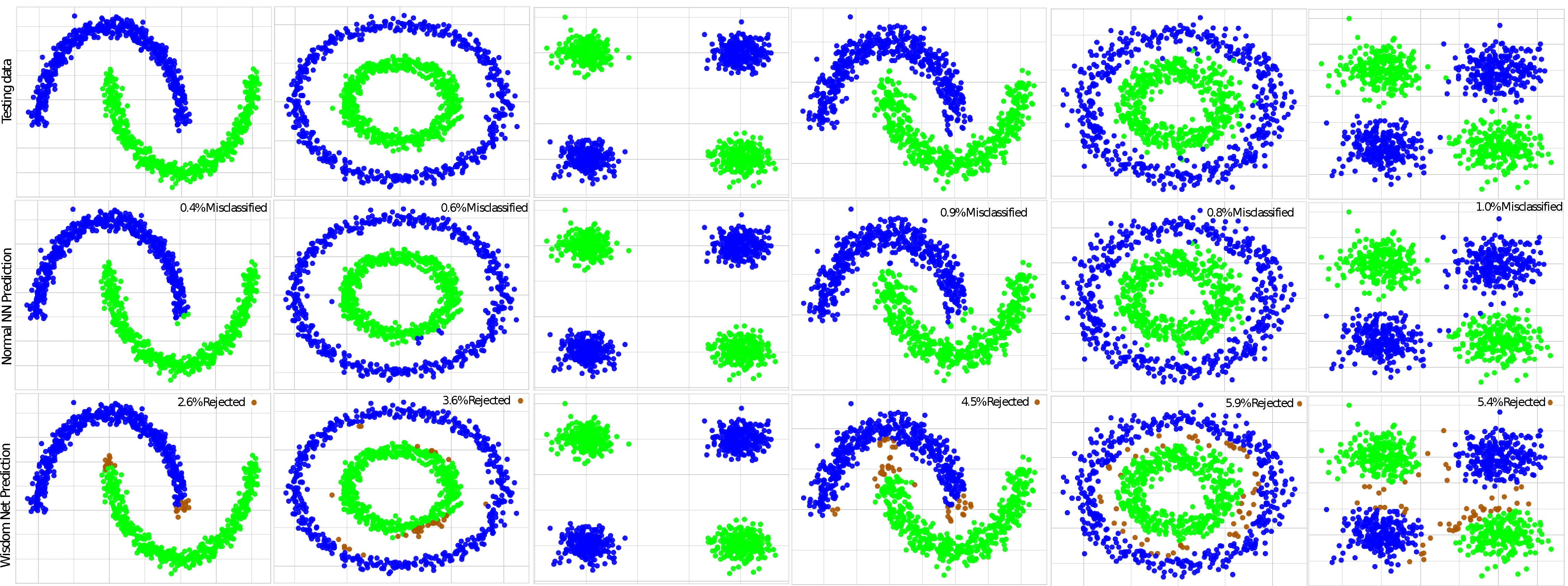}}
~~~~~~~~~~~~~~~~~~~~(a)~~~~~~~~~~~~~~~~~~~~~~~~~~~~(b)~~~~~~~~~~~~~~~~~~~~~~~~~~~~~(c)~~~~~~~~~~~~~~~~~~~~~~~~~~~~~~(d)~~~~~~~~~~~~~~~~~~~~~~~~~~~~~~(e)~~~~~~~~~~~~~~~~~~~~~~~~~~~~~~~(f)

\caption{Plot of prediction results on data with Gaussian noise. The plot containing testing data (top row), the prediction result of MLP (middle row), and the prediction result of the WisdomNet (bottom row). a) MOON with $N(\mu=0,\sigma=0.05)$, b) CIRCLE with $N(\mu=0, \sigma=0.05)$, c) XOR with $N(\mu=0, \sigma=0.05)$, d) MOON with $N(\mu=0, \sigma=0.1)$, e) CIRCLE with $N(\mu=0, \sigma=0.1)$, and f) XOR with $N(\mu=0,\sigma=0.1)$ Gaussian noise.}

\label{fig-noisy-data}
\end{figure*}

In our first experiment, noise with Gaussian distribution $(\mu=0, \sigma=0.05)$ was added to the MOON, CIRCLE, and XOR data. In the second experiment, noise with Gaussian distribution $(\mu=0, \sigma=0.1)$ was added to increase the level of noise. 
For each data set in both cases, we used 1000 training data and  1000 testing data. Fig.\ref{fig-noisy-data} shows the prediction of the WisdomNet on the testing data sets. 
The WisdomNet is able to label all misclassified samples by the MLP network as reject. By rejecting to make a decision for some data, the WisdomNet does not make any misclassification error. The reject rate is around $3\%$ for the MOON and CIRCLE data when $\sigma = 0.05$. The reject rate increases as the $\sigma$ value increases to 0.1. When $\sigma=0.1$, the reject rates for the MOON, CIRCLE, and XOR data were about $4.5\%$, $5.9\%$, and $5.4\%$, respectively. As an interesting case, when $\sigma = 0.05$ for XOR data, neither MLP nor the WisdomNet generates any prediction errors. Hence, the WisdomNet also does not yield any reject for the XOR data set in this case (Fig.\ref{fig-noisy-data}c).

We may conclude that as the noise increases, the reject rate also increases. We have tested on a higher probability of error and obtained a higher reject rate. For instance, with $\sigma =0.2$, the reject rate may increase to above $30\%$ in some test data.   

\section{Discussion} \label{discussion}
Multinomial classification is more challenging than the binary classification problem. Adding a single conjugate neuron did not provide good results in the case of multinomial classification. For example, when we classified the first 5 digits of the MNIST using a WisdomNet of one conjugate neuron, more than 50\% of the instances were rejected. A single conjugate neuron is not enough to capture all misclassifications effectively.

We have developed a WisdomNet network with multiple conjugate neurons.  For multiclass classification, we add a conjugate neuron per class that would indicate `reject' for that class only.  
This approach doubles the number of output neurons, $n_L$, in the base network yielding WisdomNet $W$ as
 $\mathcal{W}(n_{0}, n_{1}, \cdots, n_{L-1}, 2n_L)$.
In this case, all the incoming weights and biases of conjugate neurons are initially set to 0.

We tested this WisdomNet on classifying the digits 0, 1, 2, 3, and 4 of the MNIST data set. We used 30569 data for training and 5139 for testing. The base ConvNet network obtained about $99.7\%$ accuracy with $0.3\%$ error. The new structure of the WisdomNet for multinomial classification yielded a promising result. The WisdomNet reduced the misclassification error to 0 and yielded only $9.8\%$ as reject.
 For the 10-class MNIST data set, the reject rate was much higher in our experiments.  Therefore, a number of adjustments and improvements should be made to widely apply for multinomial classification problems. At this point, we conclude that the results are very promising for binary classification problems, and further studies should be conducted for multinomial classification.

An alternate idea to reject to make a decision would be to use a range of thresholds based on the predictive probabilities of the neural network for each class. 
Assume that in binary classification, the correct prediction for a data sample is expected to yield a high probability \textit{p} to a class, or low probability (\textit{1-p}) to the other class, where \textit{p} is a value between 0 and 1. When the probability outcome is around 0.5, it is hard to make a prediction. In such a case, if the predictive probability is in a range, such as in between 0.4 and 0.6, the network may reject to make a decision. We have observed that the network may misclassify a sample data by yielding a high prediction probability (e.g., $>0.95$) of belonging to an incorrect class despite the sample actually belongs to another class.  
Enlarging the range of thresholds for rejections may recognize these misclassifications; however, it would yield a significantly high ratio of rejection.

\section{Conclusion} \label{conclusion}
In this paper, we introduced trustable learning where a learning model can also learn data samples that it cannot classify and yields `reject' or `unable to classify', hence reducing the prediction error to 0. We have developed an implementation for trustable learning named WisdomNet to be applied for neural networks. We observed that this new architecture is capable of avoiding critical misclassifications and providing an opportunity for experts to study and analyze special cases. The WisdomNet is built on top of a trained base network architecture, determines the misclassified samples by the base architecture, adds one more conjugate neuron to the output layer, and then is retrained for misclassified samples using the parameters of the base network. The experimental results show that our method is generalizable since we did not restrict our method to a specific type of network architecture, activation functions, or the hyperparameters of a network.

There are many applications where error-free learning is necessary. We have tested the WisdomNet on diverse data sets with several neural network models include multilayer perceptron networks, a convolutional network, and a deep learning network. In all cases, the WisdomNet was able to reduce the error to $0\%$ and keep the reject rate under $10\%$. We evaluated the factors such as the base network accuracy, similarity of data samples, and noisy data on the reject rate. For fairly accurate neural networks, the WisdomNet is able to keep the reject rate under $5\%$. If the noise in the data increases or data is harder to classify, the reject rate increases as well. However, the reject rate still could be kept under $10\%$. It is reasonable that the reject rate in the WisdomNet is higher than the misclassification rate in the original neural networks. This result shows that the WisdomNet does not overfit the misclassified samples. In addition, when a sample is rejected by the WisdomNet, we have a chance to postpone the decision making to a human expert. Rejection to make a decision may be costly, but there could be a higher chance of making the correct prediction by a human expert. 

In the literature, whenever a learning model makes classification errors, a general method is to study the misclassified samples. Our research yields that while there are misclassified samples by a learning model, some of the samples are likely correctly classified by a chance. Our method may also be capturing those `lucky' cases and avoid potential misclassifications. 

Our experimental results are very promising if we want to achieve error-free machine learning one day. It is hard to claim that all misclassifications can be detected for any problem. However, this paper presents a big step in solving or avoiding the misclassification problem.
There are risks of generating a high reject rate or being unable to detect misclassified samples (e.g., in case of a single misclassified sample). 
If we are able to reduce the error to $0\%$, the next goal should be to reduce the reject rate as much as possible. We obtained promising results that indicate what the learning model cannot learn. For those cases, alternate learning methods can be developed to further reduce the reject rate. Furthermore, we plan to analyze strategies for multinomial classification.

In this paper, we have not discussed when to stop training for rejecting doubtful data during the fine-training process. If the fine-training continues indefinitely, there is a possibility of classifying all data as reject. Moreover, if fine-training is stopped as soon as the training error becomes zero, it may not be enough for the WisdomNet to learn the problematic testing cases. We have determined that it is possible to reach zero percent error within around $10\%$ reject rate. The optimum stopping time for fine-training is left to future work.

\section*{Conflict of interest}

The authors declare that they have no conflict of interest.

\bibliographystyle{spmpsci}
\bibliography{bibliography}

\begin{wrapfigure}{l}{0.25\columnwidth}
\begin{center}
\includegraphics[width=0.26\columnwidth]{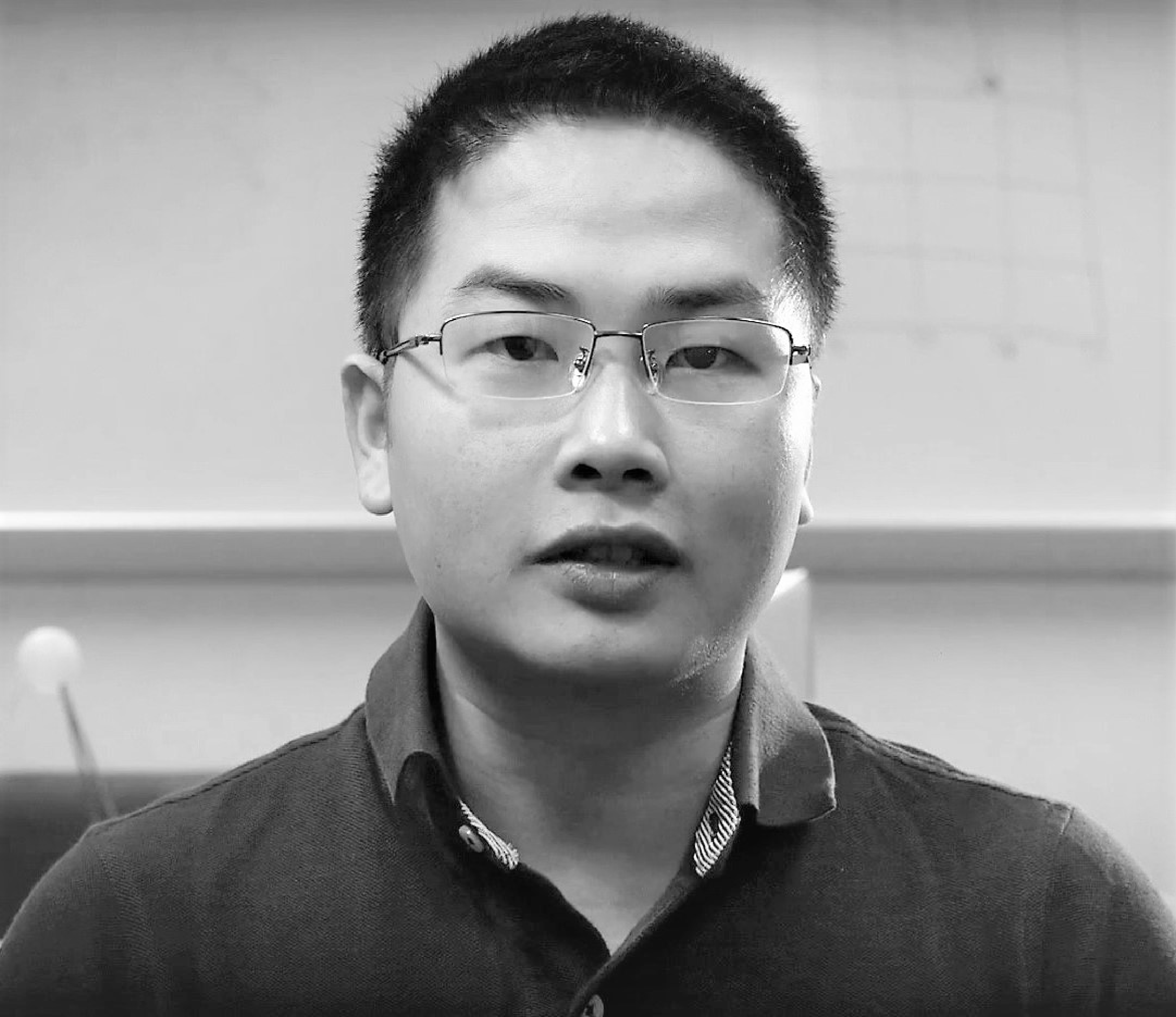}
\end{center}
\end{wrapfigure}

\textbf{Truong X. Tran} received his Bachelor's degree major in Electronic and Telecommunication Engineering, minor in Biomedical Electronic Engineering, from Hanoi University of Science and Technology, Vietnam in 2010, the M.S. degree in Computer Science from Arkansas State University, AR, USA in 2012. He is pursuing Doctoral degree in Computer Science at The University of Alabama in Huntsville, AL, USA. His research interests include machine learning and data mining.

\begin{wrapfigure}{l}{0.25\columnwidth}
\begin{center}
    \includegraphics[width=0.25\columnwidth]{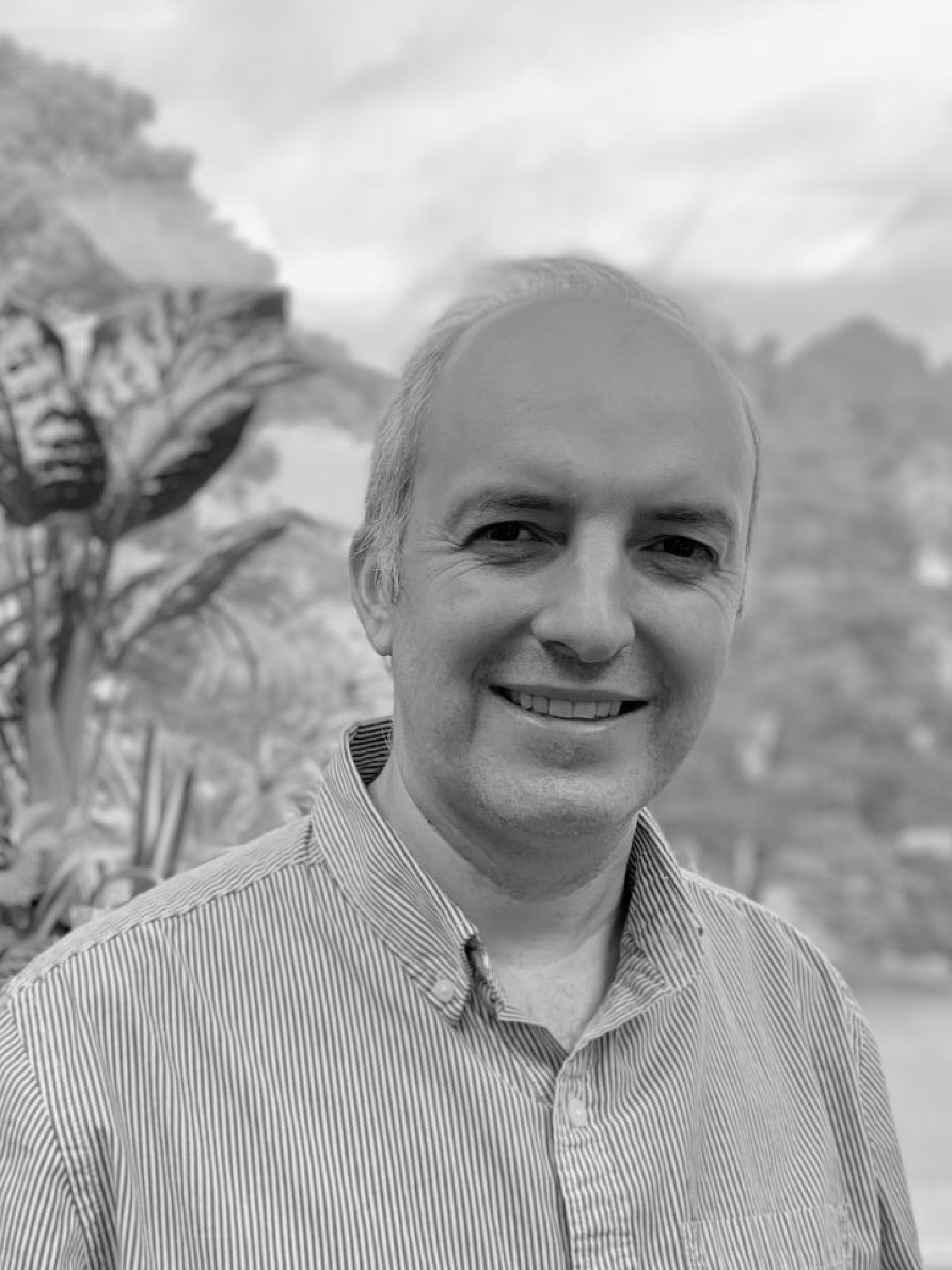}
\end{center}
\end{wrapfigure}

\textbf{Ramazan S Ayg\"{u}n:} received the B.S. degree in computer engineering from Bilkent University, Ankara, Turkey in 1996, the M.S. degree from Middle East Technical University, Ankara in 1998, and the Ph.D. degree in computer science and engineering from State University of New York at Buffalo in 2003. He is currently an Associate Professor in Computer Science Department, The University of Alabama in Huntsville. His research interests include protein crystallization image analysis, data mining, image and video processing, spatio-temporal indexing and querying, multimedia databases, semantic computing, multimedia networking, and multimedia synchronization.
\end{document}